\documentclass{article}

\usepackage{times}
\usepackage{graphicx} 
\usepackage{subfigure} 

\usepackage{natbib}

\usepackage{algorithm}
\usepackage{algorithmic}

\usepackage{helvet}
\usepackage{courier}
\usepackage{amsfonts,amssymb}
\usepackage{amsmath}
\usepackage[utf8]{inputenc} 
\usepackage[T1]{fontenc} 
\usepackage{url} 
\usepackage{booktabs} 
\usepackage{nicefrac} 
\usepackage{microtype} 

\newtheorem{theorem}{Theorem}[section]
\newtheorem{lemma}[theorem]{Lemma}

\newtheorem{corollary}[theorem]{Corollary}

\usepackage{hyperref}

\usepackage[accepted]{icml2017}

\icmltitlerunning{Asynchronous Stochastic Gradient Descent with Delay Compensation}

\begin{document} 

\twocolumn[
\icmltitle{Asynchronous Stochastic Gradient Descent with Delay Compensation}



\icmlsetsymbol{equal}{*}

\begin{icmlauthorlist}
	\icmlauthor{Shuxin Zheng}{1}
	\icmlauthor{Qi Meng}{2}
	\icmlauthor{Taifeng Wang}{3}
	\icmlauthor{Wei Chen}{3}
	\icmlauthor{Nenghai Yu}{2}
	\icmlauthor{Zhi-Ming Ma}{4} 
	\icmlauthor{Tie-Yan Liu}{3}
\end{icmlauthorlist}

\icmlaffiliation{1}{University of Science and Technology of China}
\icmlaffiliation{2}{School of Mathematical Sciences, Peking University}
\icmlaffiliation{3}{Microsoft Research}
\icmlaffiliation{4}{Academy of Mathematics and Systems Science, Chinese Academy of Sciences}

\icmlcorrespondingauthor{Shuxin Zheng}{zhengsx@mail.ustc.edu.cn}

\icmlkeywords{distributed, optimization, asynchronous, deep learning}

\vskip 0.3in
]



\printAffiliationsAndNotice{} 

\begin{abstract} 
	With the fast development of deep learning, it has become common to learn big neural networks using massive training data. Asynchronous Stochastic Gradient Descent ({ASGD}) is widely adopted to fulfill this task for its efficiency, which is, however, known to suffer from the problem of delayed gradients. That is, when a local worker adds its gradient to the global model, the global model may have been updated by other workers and this gradient becomes ``delayed''. We propose a novel technology to compensate this delay, so as to make the optimization behavior of {ASGD} closer to that of sequential {SGD}. This is achieved by leveraging Taylor expansion of the gradient function and efficient approximation to the Hessian matrix of the loss function. We call the new algorithm Delay Compensated {ASGD} ({DC-ASGD}). We evaluated the proposed algorithm on {CIFAR-10} and {ImageNet} datasets, and the experimental results demonstrate that {DC-ASGD} outperforms both synchronous {SGD} and asynchronous {SGD}, and nearly approaches the performance of sequential {SGD}.
\end{abstract} 

\section{Introduction}
Deep Neural Networks (DNN) have pushed the frontiers of many applications, such as speech recognition \cite{projection,cnnspeech}, computer vision \cite{alexnet, resnet, googlenetv4}, and natural language processing \cite{word2vec, NMT, cnnseq2seq}. Part of the success of DNN should be attributed to the availability of big training data and powerful computational resources, which allow people to learn very deep and big DNN models in parallel \cite{easgd, bmuf, ssgd-backup}.

Stochastic Gradient Descent (SGD) is a popular optimization algorithm to train neural networks \cite{sgdtricks, asgd, adam}. As for the parallelization of SGD algorithms (suppose we use $M$ machines for the parallelization), one can choose to do it in either a synchronous or asynchronous way. In synchronous SGD ({SSGD}), local workers compute the gradients over their own mini-batches of data, and then add the gradients to the global model. By using a barrier, these workers wait for each other, and will not continue their local training until the gradients from all the $M$ workers have been added to the global model. It is clear that the training speed will be dragged by the slowest worker\footnote{Recently, people proposed to use additional backup workers \cite{ssgd-backup} to tackle this problem. However, this solution requires redundant computation resources and relies on the assumption that the majority of workers train almost equally fast.}. To improve the training efficiency, asynchronous SGD (ASGD) \cite{asgd} has been adopted, with which no barrier is imposed, and each local worker continues its training process right after its gradient is added to the global model. Although ASGD can achieve faster speed due to no waiting overhead, it suffers from another problem which we call \emph{delayed gradient}. That is, before a worker wants to add its gradient $g(\textbf{w}_t)$ (calculated based on the model snapshot $\textbf{w}_t$) to the global model, several other workers may have already added their gradients and the global model has been updated to $\textbf{w}_{t+\tau}$ (here $\tau$ is called the delay factor). Adding gradient of model $\textbf{w}_t$  to another model $\textbf{w}_{t+\tau}$ does not make a mathematical sense, and the training trajectory may suffer from unexpected turbulence. This problem has been well known, and some researchers have analyzed its negative effect on the convergence speed \cite{lian2015asynchronous,avron2015revisiting}.

In this paper, we propose a novel method, called Delay Compensated ASGD (or DC-ASGD for short), to tackle the problem of delayed gradients. For this purpose, we study the Taylor expansion of the gradient function $g(\textbf{w}_{t+\tau})$ at $\textbf{w}_t$.  We find that the delayed gradient $g(\textbf{w}_t)$ is just the zero-order approximator of the correct gradient $g(\textbf{w}_{t+\tau})$, and we can leverage more items in the Taylor expansion to achieve more accurate approximation of $g(\textbf{w}_{t+\tau})$. However, this straightforward idea is practically non-trivial, because even including the first-order derivative of the gradient $g(\textbf{w}_{t+\tau})$ will require the computation of the second-order derivative of the original loss function (i.e., the Hessian matrix), which will introduce high computation and space complexity. To overcome this challenge, we propose a cheap yet effective approximator of the Hessian matrix, which can achieve a good trade-off between bias and variance of approximation, only based on previously available gradients (without the necessity of directly computing the Hessian matrix).

DC-ASGD is similar to ASGD in the sense that no worker needs to wait for others. It differs from ASGD in that it does not directly add the local gradient to the global model, but compensates the delay in the local gradient by using the approximate Taylor expansion. By doing so, it maintains almost the same efficiency as ASGD and achieves much higher accuracy. Theoretically, we proved that DC-ASGD can converge at a rate of the same order with sequential SGD for non-convex neural networks, if the delay is upper bounded; and it is more tolerant on the delay than ASGD\footnote{We also obtained similar results for the convex cases. Due to space restrictions, we put the corresponding theorems and proofs in the appendix.}. Empirically, we conducted experiments on both {CIFAR-10} and {ImageNet} datasets. The results show that (1) as compared to {SSGD} and {ASGD}, {DC-ASGD} accelerated the convergence of the training process; (2) the accuracy of the model obtained by {DC-ASGD} within the same time period is very close to the accuracy obtained by sequential SGD.

\section{Problem Setting}\label{background}
In this section, we introduce DNN and its parallel training through ASGD.

Given a multi-class classification problem, we denote $\mathcal{X}=\mathbb{R}^d$ as the input space, $\mathcal{Y}=\{1,...,K\}$ as the output space, and $\mathbb{P}$ as the joint distribution over $\mathcal{X}\times\mathcal{Y}$. Here $d$ denotes the dimension of the input space, and $K$ denotes the number of categories in the output space.

We have a training set $\{(x_1,y_1),...,(x_S,y_S)\}$, whose elements are i.i.d. sampled from $\mathcal{X}\times\mathcal{Y}$ according to distribution $\mathbb{P}$. Our goal is to learn a neural network model $O\in\mathcal{F}: \mathcal{X}\times\mathcal{Y}\rightarrow\mathbb{R}$ parameterized by $\textbf{w}\in\mathbb{R}^n$ based on the training set. Specifically, the neural network models have hierarchical structures, in which each node conducts linear combination and non-linear activation over its connected nodes in the lower layer. The parameters are the weights on the edges between two layers. The neural network model produces an output vector, i.e., $(O(x,k;\textbf{w});k\in\mathcal{Y})$ for each input $x\in\mathcal{X}$, indicating its likelihoods of belonging to different categories. Because the underlying distribution $\mathbb{P}$ is unknown, a common way of learning the model is to minimize the empirical loss function. A widely-used loss function for deep neural networks is the cross-entropy loss, which is defined as follows,
{\small\begin{equation}
	f(x,y;\textbf{w})=-\sum_{k=1}^K(I_{[y=k]}\log{\sigma_k(x;\textbf{w})}).\label{eq2.1}
	\end{equation}}Here $\sigma_k(x;\textbf{w})=\frac{e^{O(x,k;\textbf{w})}}{\sum_{k'=1}^Ke^{O(x,k';\textbf{w})}}$ is the \textit{Softmax} operator. The objective is to optimize the empirical risk, defined as below,
{\small\begin{equation}
	F(\textbf{w})=\frac{1}{S}\sum_{s=1}^Sf_s(\textbf{w}):=\frac{1}{S}\sum_{s=1}^Sf(x_s,y_s;\textbf{w}).
	\end{equation}}\begin{figure}[ht]
	\vskip 0.2in
	\begin{center}
		\centerline{\includegraphics[width=\columnwidth]{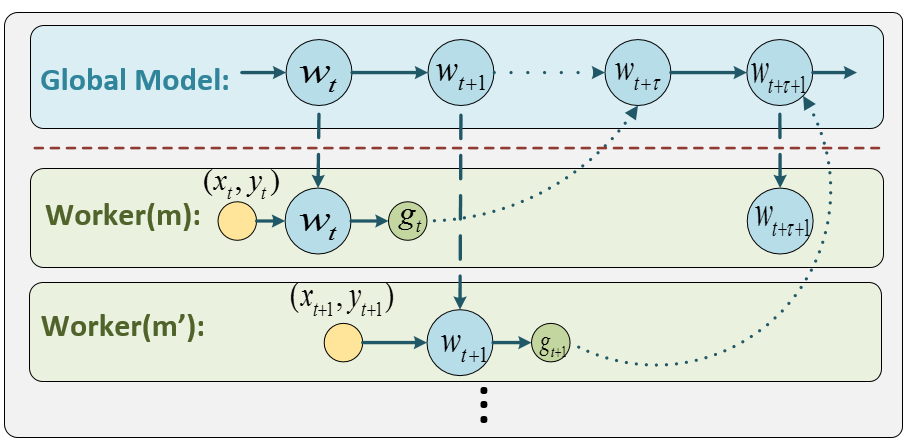}}
		\caption{ASGD training process.}
		\label{delay-sketch}
	\end{center}
	\vskip -0.2in
\end{figure} 

As mentioned in the introduction, ASGD is a widely-used approach to perform parallel training of neural networks. Although ASGD is highly efficient, it is well known to suffer from the problem of \emph{delayed gradient}. To better illustrate this problem, let us have a close look at the training process of ASGD as shown in Figure \ref{delay-sketch}. According to the figure, local worker $m$ starts from $\textbf{w}_t$, the snapshot of the global model at time $t$, calculates the local gradient $g(\textbf{w}_t)$, and then add this gradient back to the global model\footnote{Actually, the local gradient is also related to the randomly sampled data $(x_{i_t},y_{i_t})$. For simplicity, when there is no confusion, we will omit $x_{i_t},y_{i_t}$ in the notations.}. However, before this happens, some other $\tau$ workers may have already added their local gradients to the global model, the global model has been updated $\tau$ times and becomes $\textbf{w}_{t+\tau}$. The ASGD algorithm is blind to this situation, and simply adds the gradient $g(\textbf{w}_t)$ to the global model $\textbf{w}_{t+\tau}$, as follows.
\begin{equation}
\textbf{w}_{t+\tau+1}= \textbf{w}_{t+\tau}-\eta g(\textbf{w}_t),\label{eq2.2}
\end{equation}where $\eta$ is the learning rate.

It is clear that the above update rule of ASGD is problematic (and inequivalent to that of sequential SGD): one actually adds a ``delayed'' gradient $g(\textbf{w}_t)$ to the current global model $\textbf{w}_{t+\tau}$. In contrast, the correct way is to update the global model $\textbf{w}_{t+\tau}$ based on the gradient w.r.t. $\textbf{w}_{t+\tau}$. This problem of delayed gradient has been well known \cite{agarwal2011distributed,recht2011hogwild,lian2015asynchronous,avron2015revisiting}, and many practical observations indicate that it usually costs ASGD more iterations to converge than sequential SGD, and sometimes, the converged model of ASGD cannot reach accuracy parity of sequential SGD, especially when the number of workers is large \cite{asgd,ssp,easgd}. Researchers have tried to improve ASGD from different perspectives \cite{ssp,distadagrad,easgd,adadelay,mitliagkas2016asynchrony}, however, to the best of our knowledge, there is still no solution that can compensate the delayed gradient while keeping the high efficiency of ASGD. This is exactly the motivation of our paper.

\section{Delay Compensation using Taylor Expansion and Hessian Approximation}\label{theory}

As explained in the previous sections, ideally, the optimization algorithm should add gradient $g(\textbf{w}_{t+\tau})$ to the global model $\textbf{w}_{t+\tau}$, however, ASGD adds a delayed version $g(\textbf{w}_{t})$. In this section, we propose a novel method to bridge this gap by using Taylor expansion and Hessian approximation.

\subsection{Gradient Decomposition using Taylor Expansion}\label{sec3.1}

The Taylor expansion of the gradient function $g(\textbf{w}_{t+\tau})$ at $\textbf{w}_{t}$ can be written as follows \cite{folland2005higher},
{\small\begin{equation}
	g(\textbf{w}_{t+\tau})=g(\textbf{w}_{t})+\nabla g(\textbf{w}_{t})(\textbf{w}_{t+\tau}-\textbf{w}_{t})+\mathcal{O}((\textbf{w}_{t+\tau}-\textbf{w}_{t})^2)I_n,\label{eq3.1}
	\end{equation}}where $\nabla g$ denotes the matrix with the element $g_{ij}=\frac{\partial^2 f}{\partial w_i\partial w_j}$ for $i\in[n]$ and $j\in[n]$,
$(\textbf{w}_{t+\tau}-\textbf{w}_{t})^2=(w_{t+\tau,1}-w_{t,1})^{\alpha_1}\cdots(w_{t+\tau,n}-w_{t,n})^{\alpha_n}$ with $\sum_{i=1}^n\alpha_i=2$ and $\alpha_i\in\mathbb{N}$ and $I_n$ is a $n$-dimension vector with all the elements equal to $1$.

By comparing the above formula with Eqn. (\ref{eq2.2}), we can immediately find that ASGD actually uses the zero-order item in Taylor expansion as its approximation to $g(\textbf{w}_{t+\tau})$, and totally ignores all the higher-order terms $\nabla g(\textbf{w}_{t})(\textbf{w}_{t+\tau}-\textbf{w}_{t})+\mathcal{O}((\textbf{w}_{t+\tau}-\textbf{w}_{t})^2)I_n $. This is exactly the root cause of the problem of \textit{delayed gradient}. With this insight, a straightforward and ideal method is to use the full Taylor expansion to compensate the delay. However, this is practically intractable, since it involves the sum of an infinite number of items. And even the simplest delay compensation, i.e., additionally keeping the first-order item in the Taylor expansion (which is shown below), is highly non-trivial,
{\small\begin{equation}
	g(\textbf{w}_{t+\tau})\approx g(\textbf{w}_{t})+\nabla g(\textbf{w}_{t})(\textbf{w}_{t+\tau}-\textbf{w}_{t}).\label{eq3.2}
	\end{equation}}This is because the first-order derivative of the gradient function $g$ corresponds to the Hessian matrix of the original loss function $f$ (e.g., cross entropy for neural networks), which is defined as $\textbf{H}f(\textbf{w})=[h_{ij}]_{i,j=1,\cdots,n }$ where $h_{ij}=\frac{\partial^2 f}{\partial w_i \partial w_j}(\textbf{w})$.

For a neural network model with millions of parameters (which is very common and may only be regarded as a medium-size network today), the corresponding Hessian matrix will contain trillions of elements. It is clearly very computationally and spatially expensive to obtain such a large matrix\footnote{Although Hessian-free methods were used in some previous works \cite{martens2010deep}, they double the computation and communication for each local worker and are therefore not very feasible in practice.}. Fortunately, as shown in the next subsection, we find an easy-to-compute/store approximator to the Hessian matrix, which makes our proposal of delay compensation technically feasible.

\subsection{Approximation of Hessian Matrix}\label{sec3.2}

Computing the exact Hessian matrix is computationally and spatially expensive, especially for large models. Alternatively, we want to find some approximators that are theoretically close to the Hessian matrix, but can be easily stored and computed without introducing additional complexity (i.e., just using what we already have during the previous training process).

First, we show that the outer product of the gradients is an asymptotically unbiased estimation of the Hessian matrix.
Let us use $G(\textbf{w}_t)$ to denote the outer product matrix of the gradient at $\textbf{w}_t$, i.e.,
{\small\begin{equation}
	G(\textbf{w}_t)=\left(\frac{\partial}{\partial \textbf{w}}f(x,y,\textbf{w}_t)\right)\left(\frac{\partial}{\partial \textbf{w}}f(x,y,\textbf{w}_t)\right)^T.
	\end{equation}}Because the cross entropy loss is a negative log-likelihood with respect to the Softmax distribution of the model, i.e., $\mathbb{P}(Y=k|x,\textbf{w}_t)\triangleq \sigma_k (x;\textbf{w}_t)$, it is not difficult to obtain that the outer product of the gradient is an asymptotically unbiased estimation of Hessian, according to the two equivalent methods to calculate the fisher information matrix \cite{friedman2001elements}\footnote{In this paper, the norm of the matrix is Frobenius norm. }:
{\small\begin{align}\label{eq7}
	&\epsilon_t\triangleq\mathbb{E}_{(y|x,\textbf{w}^*)}|| G(\textbf{w}_t)- H(\textbf{w}_t)|| \to 0 , t\to\infty.
	\end{align}}The assumption behind the above equivalence is that the underlying distribution equals the model distribution with parameter $\textbf{w}^*$ (or there is no approximation error of the NN hypothesis space) and the training model $\textbf{w}_t$ gradually converges to the optimal model $\textbf{w}^*$ along with the training process. This assumption is reasonable considering the universal approximation property of DNN \cite{hornik1991approximation} and the recent results on the optimality of the local optima of DNN \cite{choromanska2015loss, kawaguchi2016deep}.

Second, we show that by further introducing a well-designed weight to the outer product of the gradients, we can achieve a better trade-off between bias and variance for the approximation.

Although the outer product of the gradients can achieve unbiased estimation to the Hessian matrix, it may induce high approximation error due to potentially large variance.
To further control the variance, we use mean square error (MSE) to measure the quality of an approximator, which is defined as follows,
{\small
	\begin{equation}
	mse^t(G)= \mathbb{E}_{(y|x,\textbf{w}^*)}\|\big(G(\textbf{w}_t)- H(\textbf{w}_t)\big)||^2.
	\end{equation}
	We consider the following
	new approximator {\small$\lambda G(\textbf{w}_t)\overset{\Delta}{=}\left[\lambda g_{ij}^t\right]$}, and prove that with appropriately set $\lambda$,
	$\lambda G(\textbf{w}_t) $ can lead to smaller MSE than $ G(\textbf{w}_t) $, for arbitrary model $\textbf{w}_t$ during the training. 	
	
	\begin{theorem}\label{thm3.2}
		Assume that the loss function is $L_1$-Lipschitz, and for arbitrary $k\in[K]$,   $\left|\frac{\partial \sigma_k}{\partial w_i}\right|\in[l_{i},u_{i}]$,  $|\frac{\sigma_k(x, \textbf{w}^*)}{\sigma_k(x, \textbf{w}_t)}|\in[\alpha,\beta]$. If {\small$\lambda\in [0, 1]$} makes the following inequality holds, {\small\begin{equation}\label{eq9}\sum_{k=1}^{K}\frac{1}{\sigma_k^3(x,\textbf{w}_t)}\geq2C\left[\left(\sum_{k=1}^{K}\frac{1}{\sigma_k(x,\textbf{w}_t)}\right)^2+2L_1^2\epsilon_t\right],\end{equation}} where $C=\max_{i,j} \frac{1}{1+\lambda}(\frac{u_iu_j\beta}{l_il_j\alpha})^2$, and the model $\textbf{w}_t$ converges to the optimal model $\textbf{w}^*$, then $mse^t(\lambda G)\leq mse^t (G)$.
	\end{theorem}
	
	The following corollary gives simpler sufficient conditions for Theorem \ref{thm3.2}.
	
	\begin{corollary}\label{coro3.3}
		A sufficient condition for inequality (\ref{eq9}) is $\exists k_0\in[K]$ such that {\small$\sigma_{k_0}\in\left[1-\frac{K-1}{2C(K^2+L_1^2\epsilon_t)},1\right]$}.
	\end{corollary}
	
	According to Corollary \ref{coro3.3}, we have the following discussions. Please note that, if $\textbf{w}_t$ converges to $\textbf{w}^*$, $\epsilon_t$ is a decreasing term and approaches $0$. Thus, $\epsilon_t$ can be upper bounded by a very small constant for large $t$. Therefore, the condition on $\sigma_k(x,\textbf{w}_t)$ is more likely to be satisfied when $\sigma_k(x,\textbf{w}_t)$  ($\exists k\in[K]$) is close to $1$. Please note that this is not a strong condition, since if $\sigma_k(x,\textbf{w}_t)$ ($\forall k\in[K]$) is very small, the classification power of the corresponding neural network model will be very weak and not useful in practice.
	
	Third, to reduce the storage of the approximator {\small$\lambda G(\textbf{w})$},  we adopt a widely-used diagonalization trick \cite{diagH}, which has shown promising empirical results. To be specific, we only store the diagonal elements of the approximator $\lambda G(\textbf{w})$ and make all the other elements to be zero. We denote the refined approximator as $Diag(\lambda G(\textbf{w}))$ and assume that the diagonalization error is upper bounded by $\epsilon_D$, i.e., $||Diag(H(\textbf{w}_t))- H(\textbf{w}_t)||\leq \epsilon_D$. We give a uniform upper bound of its MSE in the supplementary materials, from which we can see that $\lambda$ plays a role of trading off variance and Lipschitz\footnote{See \textbf{Lemma 3.1} in Supplementary.}.

	\section{\textit{Delay Compensated ASGD}: Algorithm Description}\label{sec4}
	In Section \ref{theory}, we have shown that {\small$Diag(\lambda G(\textbf{w}))$} is a cheap approximator of the Hessian matrix, with guaranteed approximation accuracy. In this section, we will use this approximator to compensate the gradient delay, and call the corresponding algorithm Delay-Compensated ASGD (DC-ASGD). Since $ Diag(\lambda G(\textbf{w}))=\lambda g(\textbf{w}_t)\odot g(\textbf{w}_t)$, where $\odot$ indicates the element-wise product, the update rule for DC-ASGD can be written as follows:
	{\small\begin{align}\label{eq4.1}
		\textbf{w}_{t+\tau+1}= \textbf{w}_{t+\tau}- \eta \left(g(\textbf{w}_t) + \lambda g(\textbf{w}_t)\odot g(\textbf{w}_t) \odot ({\textbf{w}}_{t+\tau}-\textbf{w}_t) \right),
		\end{align}
	}We call $ g(\textbf{w}_t) + \lambda g(\textbf{w}_t)\odot g(\textbf{w}_t) \odot ({\textbf{w}}_{t+\tau}-\textbf{w}_t)$ the \emph{delay-compensated gradient} for ease of reference.
	
	\begin{algorithm}[tb]
		\caption{DC-ASGD: worker $m$}
		\label{alg1}
		\begin{algorithmic}
			\REPEAT
			\STATE Pull $\textbf{w}_t$ from the parameter server.
			\STATE Compute gradient $g_m = \nabla f_m(\textbf{w}_t)$.
			\STATE Push $g_m$ to the parameter server.
			\UNTIL{$forever$}
		\end{algorithmic}
	\end{algorithm}
	
	\begin{algorithm}[tb]
		\caption{DC-ASGD: parameter server}
		\label{alg2}
		\begin{algorithmic}
			\STATE {\bfseries Input:} learning rate $\eta$, variance control parameter $\lambda_t$.
			\STATE {\bfseries Initialize:} $t=0$, $\textbf{w}_0$ is initialized randomly, $\textbf{w}_{bak}(m) = \textbf{w}_0$, $m \in \{1,2,\cdots,M\}$
			\REPEAT
			\IF{receive ``$g_m$"} 
			\STATE ${\textbf{w}}_{t+1} \leftarrow {\textbf{w}}_{t} - \eta \cdot \big( g_m + \lambda_tg_m\odot g_m\odot ({\textbf{w}}_{t}-\textbf{w}_{bak}(m))\big)$
			\STATE $t\leftarrow t+1$
			\ELSIF{receive ``pull request''}
			\STATE $\textbf{w}_{bak}(m) \leftarrow {\textbf{w}}_{t}$
			\STATE Send ${\textbf{w}}_{t}$ back to worker $m$.
			\ENDIF
			\UNTIL{$forever$}
		\end{algorithmic}
	\end{algorithm}
	
	The flow of DC-ASGD is shown in Algorithms \ref{alg1} and \ref{alg2}. Here we assume that DC-ASGD is implemented by using the parameter server framework (although it can also be implemented in other frameworks). According to Algorithm \ref{alg1}, local worker $m$ pulls the latest global model $\textbf{w}_t$ from the parameter server, computes its gradient $g_m$ and sends it back to the server. According to Algorithm \ref{alg2}, the parameter server will store a backup model $\textbf{w}_{bak}(m)$ when worker $m$ pulls $\textbf{w}_{t}$. When the delayed gradient $g_m$ calculated by worker $m$ is received at time $t$, the parameter server updates the global model according to Eqn (\ref{eq4.1}).
	
	Please note that as compared to ASGD, DC-ASGD has no extra communication cost and no extra computational requirement on the local workers. And the additional computations regarding Eqn(\ref{eq4.1}) only introduce a lightweight overhead to the parameter server. As for the space requirement, for each worker $m \in \{1,2,\cdots,M\}$, the parameter server needs to additionally store a backup model $\textbf{w}_{bak}(m)$. This is not a critical issue since the parameter server is usually implemented in a distributed manner, and the parameters and its backup version are stored in CPU-side memory which is usually far beyond the total parameter size. In this case, the cost of DC-ASGD is quite similar to ASGD, which is also reflected by our experiments.
	
	The Delay Compensation is not only applicable to ASGD but SSGD. Recently a study on SSGD\cite{fbimagenet} assumes $g(\textbf{w}_{t+j})\approx g(\textbf{w}_{t})$ for $j<M$ to make the updates from small and large mini-batch SGD similar, which can be immediately improved by applying delay-compensated gradient. Please check the detailed discussion in Supplementary.

	\section{Convergence Analysis}\label{3.3}
	In this section, we prove the convergence rate of DC-ASGD. Due to space restrictions, we only give the results for the non-convex case, and leave the results for the convex case (which is much easier) to the supplementary.
	
	In order to present our main theorem, we need to introduce the following mild assumptions.

	\textbf{Assumption 1 (Smoothness):} \cite{lian2015asynchronous}\cite{recht2011hogwild}
	The loss function is smooth w.r.t. the model parameter, and we use $L_1,L_2,L_3$ to denote the upper bounds of the first, second, and third-order derivatives of the loss function.  The activation function $\sigma_k(\textbf{w})$ is $L$-Lipschitz continuous.
	
	\textbf{Assumption 2 (Non-convexity):}  \cite{lee2016gradient} The loss function is $\mu$-strongly convex in a ball centered at each local optimum  which is denoted as $d(\textbf{w}_{loc},r)$ with radius $r$, and twice differential about $\textbf{w}$.

	We also introduce some notations to simplify the presentation of our results, i.e., {\small\begin{equation*}
		M=\max_{k, \textbf{w}_{loc}}\left|\mathbb{P}(Y=k|x,\textbf{w}_{loc})-\mathbb{P}(Y=k|x,\textbf{w}^*)\right|,
		\end{equation*}}
	{\small\begin{equation*}
		\begin{split}
		H=\max_{k,x,\textbf{w}}\left|\frac{\partial^2 \mathbb{P}(Y=k|x,\textbf{w})}{\partial^2\textbf{w}}\times\frac{1}{\mathbb{P}(Y=k|x,\textbf{w})}\right|,\\ 
		\forall k\in [K],x,w.
		\end{split}
		\end{equation*}}

	Actually, the non-convexity error $\epsilon_{nc}=HKM$, which is defined as the upper bound of the difference between the prediction outputs of the local optima and the global optimum (Please see Lemma 5.1 in the supplementary materials). We assume that the DC-ASGD search in the set $\|\textbf{w}-\textbf{w}'\|_2^2\leq \pi^2, \forall \textbf{w}, \textbf{w}'$ and denote {\small $D_0=F(\textbf{w}_1)-F(\textbf{w}^*)$},  {\small$C_{\lambda}^2=(L_3^2\pi^2/2+2((1-\lambda)L_1^2+\epsilon_D)^2+2\epsilon_{nc}^2)$}, $\tilde{C}_{\lambda}^2=4T_0\max_{s=1,\cdots,T_0}{\epsilon_s}^2+4\theta^2\log{(T-T_0)}$ where $T_0\geq \mathcal{O}(1/r^4)$, $\theta=\frac{2HKLVL_2}{\mu^2}\sqrt{\frac{1}{\mu}\left(1+\frac{L_2+\lambda L_1^2}{L_2}\tau\right)}$.
	
	With all the above, we have the following theorem.
	
	\begin{theorem}\label{thm3.8}
		Assume that Assumptions 1-2 hold. Set the learning rate
		{\small$
			\eta=\sqrt{\frac{2D_0}{bTL_2V^2}},
			$}where $b$ is the mini-batch size, and $V$ is the upper bound of the variance of the delay-compensated gradient. If $T\geq \max\{\mathcal{O}(1/r^{4}), 2D_0bL_2/V^2\}$ and delay $\tau$ is upper-bounded as below,
		{\small\begin{align}
			&\tau\leq\min\left\{\frac{L_2\gamma}{C_{\lambda}},\frac{\gamma}{C_{\lambda}},\frac{\sqrt{T}\gamma}{\tilde{C}},\frac{L_2T\gamma}{4\tilde{C}}\right\},
			\end{align}} where $\gamma=\sqrt{\frac{L_2 TV^2}{2D_0b}}$, then DC-ASGD has the following ergodic convergence rate,
		\begin{equation}
		\min_{t=\{1,\cdots,T\}}\mathbb{E}(\|\nabla F(\textbf{w}_t)\|^2)\leq V\sqrt{\frac{2D_0L_2}{bT}},
		\end{equation}
		where $T$ is the number of iteration, the expectation is taken with respect to the random sampling in  SGD and the data distribution $P(Y|x,\textbf{w}^*)$.
	\end{theorem}
	
	\textit{Proof Sketch\footnote{Please check the complete proof in the supplementary material.}:}
	
	\textit{Step 1:} We denote the delay-compensated gradient as $g_m^{dc}(w_{t})$ where $m\in\{1,\cdots,b\}$ is the index of instances in the mini-batch and $\nabla F^{h}(w_t)=\nabla F(w_t)+\mathbb{E}H(w_t)(w_{t+\tau}-w_t)$. According to Assumption 1, we have
	{\small\begin{eqnarray}\nonumber
		&&\mathbb{E}F(w_{t+\tau+1})-F(w_{t+\tau})\\\nonumber
		&\leq&-\frac{b\eta_{t+\tau}}{2}\left(\|\nabla F(w_{t+\tau})\|^2+\left\|\sum_{m=1}^b\mathbb{E}g_m^{dc}(w_t)\right\|^2\right)\\\nonumber
		&&+b\eta_{t+\tau}\left\|\nabla F(w_{t+\tau})-\sum_{m=1}^b\nabla F^{h}(w_t)\right\|^2\\
		&&+b\eta_{t+\tau}\left\|\sum_{m=1}^b\mathbb{E}g_m^{dc}(w_t)-\sum_{m=1}^bF^{h}(w_t)\right\|^2 \nonumber\\
		&&+\frac{\eta_{t+\tau}^2L_2}{2}\mathbb{E}\left(\left\|\sum_{m=1}^bg_m^{dc}(w_t)\right\|^2\right).\label{eq2}
		\end{eqnarray}}The term {\small$\left\|\sum_{m=1}^b\mathbb{E}g_m^{dc}(w_t)-\sum_{m=1}^bF^{h}(w_t)\right\|^2$}, measured by the expectation with respect to $\mathbb{P}(Y|x,w^*)$, is bounded by $C_{\lambda}^2\cdot\|w_{t+\tau}-w_t\|^2$. The term {\small$ \left\|\nabla F(w_{t+\tau})-\sum_{m=1}^b\nabla F^{h}(w_t)\right\|^2$} can be bounded by $\frac{L_3^2}{4}\|w_{t+\tau}-w_t\|^4$, which will be smaller than $\|w_{t+\tau}-w_t\|^2$ when $\|w_{t+\tau}-w_t\|$ is small. Other terms which are related to the gradients can be further upper bounded by the smoothness property of the loss function.
	
	\textit{Step 2:} We proved that, under the non-convexity assumption, if $\|\lambda g(w_t)\odot g(w_t)\|\leq \lambda L_1^2$, then when $t>\mathcal{O}(1/r^4)$, $\epsilon_t\leq\theta\sqrt{\frac{1}{t-T_0}}+\epsilon_{nc}$, where $T_0=\mathcal{O}(1/r^4)$. That is, we can find a weaker condition for the decreasing of $\epsilon_t$ than that for $\textbf{w}_t\to\textbf{w}^*$.
	
	\textit{Step 3:} By plugging in the decreasing rate of $\epsilon_t$ in \textit{Step 1} and following a similar proof of the convergence rate of ASGD \cite{lian2015asynchronous}, we can get the result in the theorem.
	
	\begin{figure*}[t]
		\centering
		\includegraphics[scale=0.34]{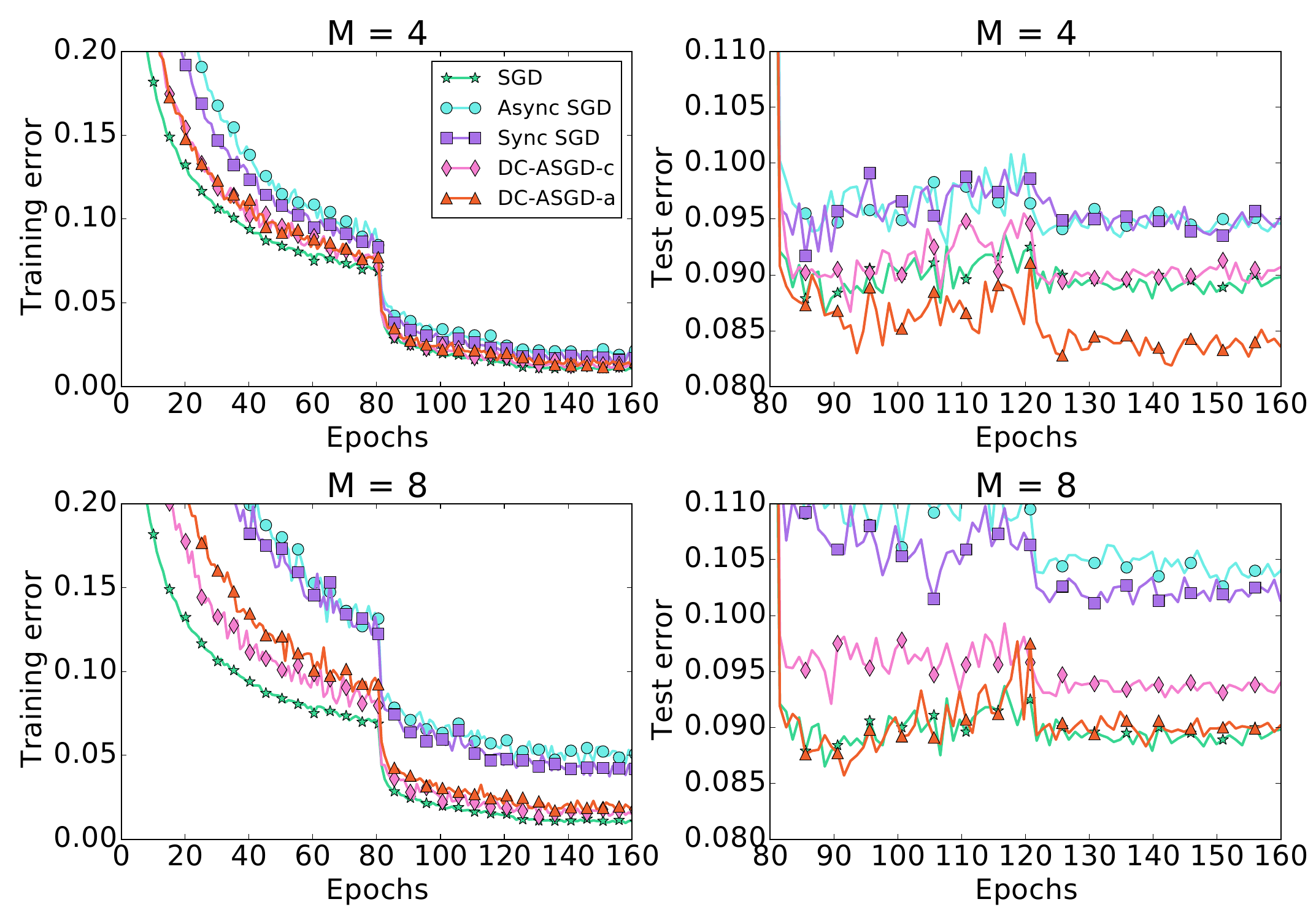}
		\caption{Error rates of the global model w.r.t. number of effective passes of data on CIFAR-10}
		\label{cifar10}
	\end{figure*}

	\textbf{Discussions:}
	
	(1) The above theorem shows that the convergence rate of DC-ASGD is in the order of $O(\frac{V}{\sqrt{T}})$. Recall that the convergence rate of ASGD is $O(\frac{V_1}{\sqrt{T}})$, where $V_1$ is the variance for the delayed gradient $g(w_t)$. By simple calculation, $V$ can be upper bounded by $V_1+\lambda V_2$, where $V_2$ is the extra moments of the noise introduced by the delay compensation term. Thus if we set $\lambda\in [0, V_1/V_2]$, DC-ASGD and ASGD will converge at the same rate.
	As the training process goes on, $g(w)$ will become smaller. Compared with $V_1$, $V_2$ (composed by variance of $g\odot g$) will not be the dominant order and can be gradually neglected. As a result, the feasible range for $\lambda$ is actually very large.
	
	(2) Although DC-ASGD converges at the same rate with ASGD, its tolerance on the delay is much better if
	$T\geq\max\{\tilde{C}^2,4\tilde{C}/L_2\}$ and $C_{\lambda}<\min\{L_2,1\}$. The intuition for the condition on $T$ is that larger $T$ induces smaller step size $\eta$. A small step size means that $w_t$ and $w_{t+\tau}$ are close to each other. According to the upper bound of Taylor expansion series \cite{folland2005higher}, we can see that \emph{delay compensated gradient} will be more accurate than the delayed gradient used in ASGD. Since $C_{\lambda}$ is related to the diagonalization error $\epsilon_D$ and the non-convexity error $\epsilon_{nc}$, smaller $\epsilon_D $ and $\epsilon_{nc}$ will lead to looser conditions for the convergence. If these two error are sufficiently small (which is usually the case according to \cite{choromanska2015loss,kawaguchi2016deep, yann1987modeles}), the condition $L_2>C_{\lambda}$ can be simplified as $L_2>(1-\lambda)L_1^2+L_3\pi$, which is easy to be satisfied with a small $1-\lambda$. Assume that $L_2-L_3\pi>0$, which is easily to be satisfied if the gradient is small (e.g. at the later stage of the training progress). Accordingly, we can obtain the feasible range for $\lambda$ as $\lambda\in[1-(L_2-L_3\pi)/2L_1^2,1]$. $\lambda$ can be regarded as a trade-off between the extra variance introduced by the delay-compensate term $\lambda g(w_t)\odot g(w_t)$ and the bias in Hessian approximation.
	
	(3) Actually ASGD is an extreme case for DC-ASGD, with $\lambda=0$. Another extreme case is with $\lambda=1$. 
	DC-ASGD prefers larger $T$ and smaller $\pi$, which can lead to a faster speed-up and larger tolerant for delay.
	
	Based on the above discussions, we have the following corollary, which indicates that DC-ASGD is superior to ASGD in most cases.
	\begin{corollary}
		Let $C_0=\max\{\tilde{C}^2,4\tilde{C}/L_2\}$, which is a constant. If we choose $\lambda\in \left[1-\frac{L_2-L_3\pi}{L_1^2},1\right]\cap [0,V_1/V_2]\cap [0,1]$ and the number of total iterations $T\geq C_0$, DC-ASGD will outperform ASGD by a factor of $T/C_0$.
	\end{corollary}

	\section{Experiments}\label{experiment}
	
	In this section, we evaluate our proposed DC-ASGD algorithm. We used two datasets: CIFAR-10 \cite{cifar10} and ImageNet ILSVRC 2013 \cite{imagenet}. The experiments were conducted on a GPU cluster interconnected with InfiniBand. Each node has four K40 Tesla GPU processors. We treat each GPU as a separate local worker. For the DNN algorithm running on each worker, we chose ResNet \cite{resnet} since it produces the state-of-the-art accuracy in many image related tasks and its implementation is available through open-source projects\footnote{\url{https://github.com/KaimingHe/deep-residual-networks}}. For the parallelization of ResNet across machines, we leveraged an open-source parameter server\footnote{\url{http://www.dmtk.io/}}.
	
	We implemented DC-ASGD on this experimental platform. We have two versions of implementations, one sets $\lambda_t = \lambda_0$ as a constant, and the other adaptively tunes $\lambda_t$ using a moving average method proposed by \cite{rmsprop}. Specifically, we first define a quantity called MeanSquare as follows, {\small\begin{equation}\label{eq18}
		MeanSquare(t) = m \cdot MeanSquare(t-1) + (1-m) \cdot g(\textbf{w}_t)^2,
		\end{equation}
	}where $m$ is a constant taking value from $[0,1)$. And then we divide the initial $\lambda_0$ by $\sqrt{MeanSquare(t) + \epsilon}$, where $\epsilon = 10^{-7}$ for all our experiments. This adaptive method is adopted to reduce the variance among coordinates with historical gradient values. For ease of reference, we denote the first implementation as DC-ASGD-c (constant) and the second as DC-ASGD-a (adaptive).
	\begin{figure*}
		\centering
		\includegraphics[scale=0.34]{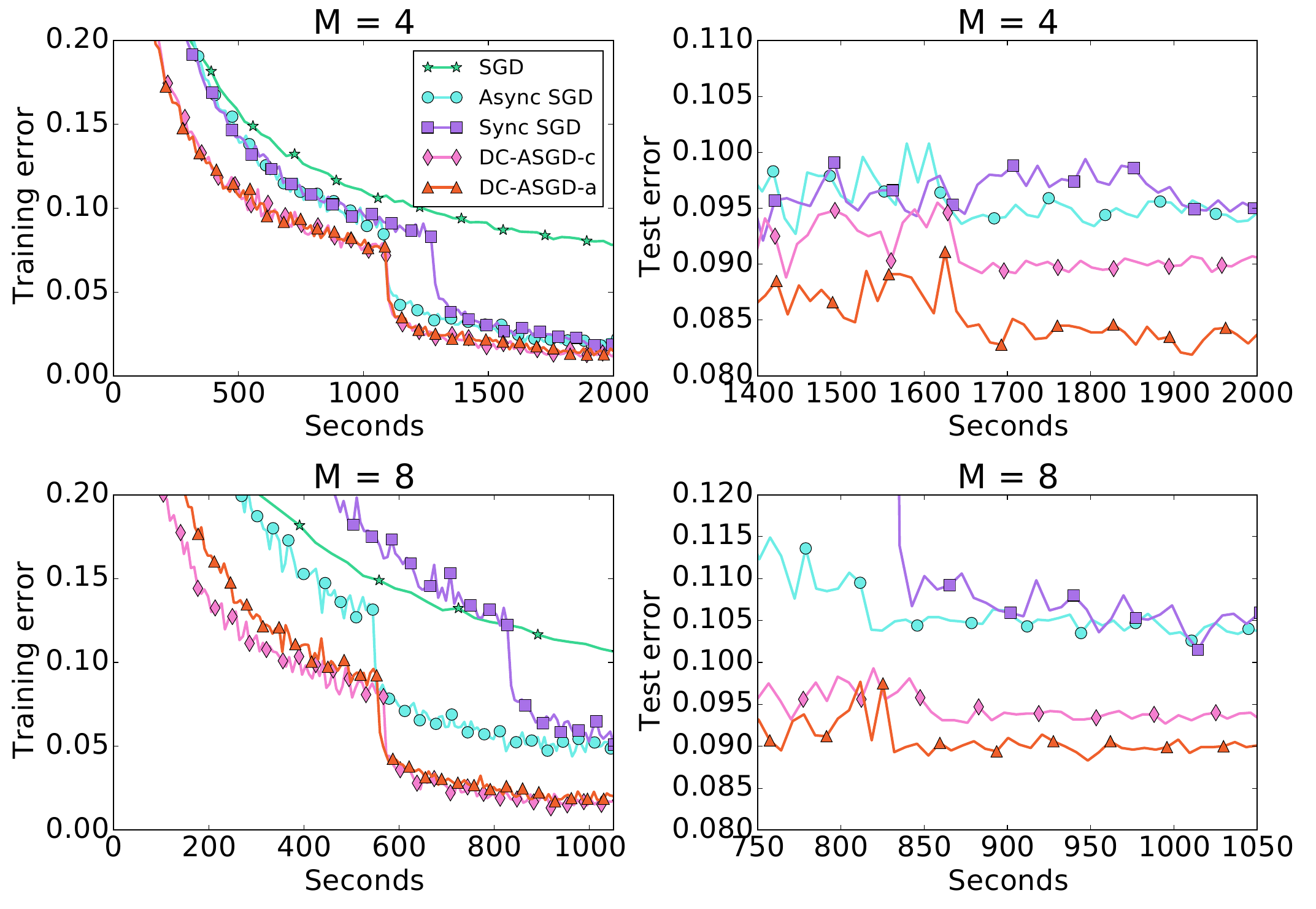}
		\caption{Error rates of the global model w.r.t. wallclock time on CIFAR-10}
		\label{cifar10_p}
	\end{figure*}
	
	In addition to DC-ASGD, we also implemented ASGD and SSGD, which have been used in many previous works as baselines \cite{asgd, ssgd-backup, das2016distributed}. Furthermore, for the experiments on CIFAR-10, we used the sequential SGD algorithm as a reference model to examine the accuracy of parallel algorithms. However, for the experiments on ImageNet, we were not able to show this reference because it simply took too long time for a single machine to finish the training\footnote{We also implemented the momentum variants of these algorithms. The corresponding comparisons are very similar to those without momentum.}. For sake of fairness, all experiments started from the same randomly initialized model, and used the same strategy for learning rate scheduling. The data were repartitioned randomly onto the local workers every epoch.

	\subsection{Experimental Results on CIFAR-10}\label{cifar10exp}
	
	The CIFAR-10 dataset consists of a training set of 50k images and a test set of 10k images in 10 classes. We trained a 20-layer ResNet model on this dataset (without data augmentation). For all the algorithms under investigation, we performed training for 160 epochs, with a mini-batch size of 128, and an initial learning rate which was reduced by ten times after 80 and 120 epochs following the practice in \cite{resnet}. We performed grid search for the hyper-parameter and the best test performances are obtained by choosing the initial learning rate $\eta=0.5$, $\lambda_0=0.04$ for DC-ASGD-c, and $\lambda_0=2$, $m=0.95$ for DC-ASGD-a. We tried different numbers of local workers in our experiments: $M=\{1,4,8\}$.

	\begin{table}[ht]
		\renewcommand\arraystretch{1.1}				\setlength{\tabcolsep}{8pt}
		\caption{Classification error on CIFAR-10 test set. The number of $\dagger$ is 8.75 reported in \cite{resnet}. Fig.~\ref{cifar10} and \ref{cifar10_p} show the training procedures.}
		\begin{center}
			\small
			\begin{tabular}{c|c|c}
				\hline
				\# workers& algorithm & error(\%) \\
				\hline
				1 & SGD & 8.65$^\dagger$ \\
				\hline
				4 & ASGD & 9.27 \\
				& SSGD & 9.17 \\
				& DC-ASGD-c & 8.67\\
				& DC-ASGD-a & \textbf{8.19}\\
				\hline
				8 & ASGD & 10.26\\
				& SSGD & 10.10\\
				& DC-ASGD-c & 9.27\\
				& DC-ASGD-a & \textbf{8.57}\\
				\hline
			\end{tabular}
		\end{center}
		\vspace{-1em}
		\label{tab:cifar10}
	\end{table}
	
	First, we investigate the learning curves with fixed number of effective passes as shown in Figure \ref{cifar10}. From the figure, we have the following observations: (1) Sequential SGD achieves the best accuracy, and its final test error is 8.65\%. (2) The test errors of ASGD and SSGD increase with respect to the number of local workers. In particular, when $M = 4$, ASGD and SSGD achieve test errors of 9.27\% and 9.17\% respectively; and when $M=8$, their test errors become 10.26\% and 10.10\% respectively. These results are reasonable: ASGD suffers from delayed gradients which becomes more serious for a larger number of workers; SSGD increases the effective mini-batch size by $M$ times, and enlarged mini-batch size usually affects the training performances of DNN. (3) For DC-ASGD, no matter which $\lambda_t$ is used, its performance is significantly better than ASGD and SSGD, and catches up with sequential SGD. For example, when $M=4$, the test error of DC-ASGD-c is 8.67\%, which is indistinguishable from sequential SGD, and the test error for DC-ASGD-a is 8.19\%, which is even better than that achieved by sequential SGD. It is not by design that DC-ASGD can beat sequential SGD. The test performance lift might be attributed to the regularization effect brought by the variance introduced by parallel training. When $M=8$, DC-ASGD-c can reduce the test error to 9.27\%, which is nearly 1\% better than ASGD and SSGD, meanwhile the test error is 8.57\% for DC-ASGD-a, which again slightly better than sequential SGD.
	\begin{figure*}
		\centering
		\includegraphics[scale=0.34]{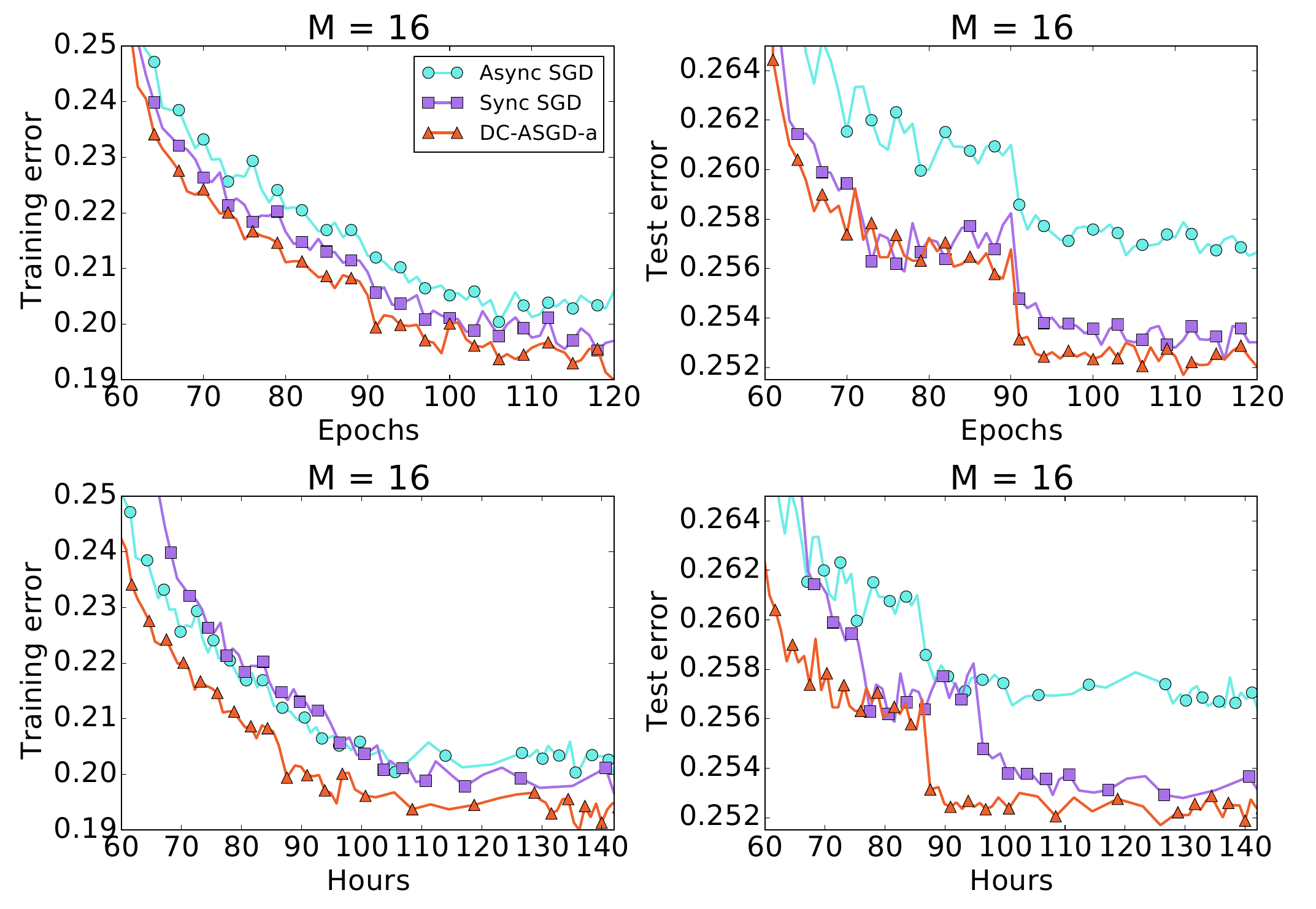}
		\caption{Error rates of the global model w.r.t. both number of effective passes and wallclock time on ImageNet}
		\label{imagenet}
	\end{figure*}

	We further compared the convergence speeds of different algorithms as shown in Figure \ref{cifar10_p}. From this figure, we have the following observations: (1) Although the convergent point is not very good, ASGD runs indeed very fast, and achieves almost linear speed-up as compared to sequential SGD in terms of throughput. (2) SSGD also runs faster than sequential SGD. However, due to the synchronization barrier, it is significantly slower than ASGD. (3) DC-ASGD achieves very good balance between accuracy and speed. On one hand, its converge speed is very similar to that of ASGD (although it involves a little more computational cost and some memory cost when compensating the delay). On the other hand, its convergent point is as good as, or even better than that of sequential SGD. The experiments results clearly demonstrate the effectiveness of our proposed delay compensation technologies\footnote {Please refer to the supplementary materials for the experiments on tuning the parameter $\lambda$.}.
	
	\subsection{Experimental Results on ImageNet }

	In order to further verify our method on the large-scale setting, we conducted the experiment on the ImageNet dataset, which contains 1.28 million training images and 50k validation images in 1000 categories. We trained a 50-layer ResNet model \cite{resnet} on this dataset.
	
	According to the previous subsection, DC-ASGD-a seems to be better, therefore in this large-scale experiment, we only implemented DC-ASGD-a.  For all algorithms in this experiment, we performed training for 120 epochs , with a mini-batch size of 32, and an initial learning rate reduced by ten times after every 30 epochs following the practice in \cite{resnet}.
	We did grid search for hyperparameter tuning and set the initial learning rate $\eta=0.1$, $\lambda_0=2$, $m=0$. Since the training on the ImageNet dataset is very time consuming, we employed $M=16$ GPU nodes in our experiments. The top-1 accuracies based on \textbf{1-crop} testing
	of different algorithms are given in Figure \ref{imagenet}.

	\begin{table}[ht]
		\vspace{-.5em}
		\renewcommand\arraystretch{1.1}
		\setlength{\tabcolsep}{8pt}
		\caption{Top-1 error on \textbf{1-crop} ImageNet validation. Fig.~\ref{imagenet} shows the training procedures.}
		\begin{center}
			\small
			\begin{tabular}{c|c|c}
				\hline
				\# workers& algorithm & error(\%) \\
				\hline
				16 & ASGD & 25.64\\
				& SSGD & 25.30\\
				& DC-ASGD-a & \textbf{25.18}\\
				\hline
			\end{tabular}
		\end{center}
		\vspace{-0em}
		\label{tab:imagenet}
	\end{table}
	
	According to the figure, we have the following observations: (1) After processing the same amount of training data, DC-ASGD always outperforms SSGD and ASGD. In particular, while the eventual test error achieved by ASGD and SSGD were 25.64\% and 25.30\% respectively, DC-ASGD achieved a lower error rate of 25.18\%. Please note this time the accuracy of SSGD is quite good (which is consistent with a separate observation in \cite{ssgd-backup}). An explanation is that the training on ImageNet is less sensitive to the mini-batch size than that on CIFAR-10. (2) If we look at the learning curve with respect to wallclock time, SSGD is slowed down due to the synchronization barrier; ASGD and DC-ASGD have similar efficiency, once again indicating that the extra overhead for delay compensation introduced by DC-ASGD can almost be neglected in practice. 
	Based on all our experiments, we can clearly see that DC-ASGD has outstanding performance in terms of both classification accuracy and convergence speed, which in return verifies the soundness of our proposed delay compensation technologies.

	\section{Conclusion}\label{conclusion}
	In this paper, we have given a theoretical analysis on the problem of delayed gradients in the asynchronous parallelization of stochastic gradient descent (SGD) algorithms, and proposed a novel algorithm called Delay Compensated Asynchronous SGD (DC-ASGD) to tackle the problem. We have evaluated DC-ASGD on CIFAR-10 and ImageNet datasets, and the results demonstrate that it can achieve better accuracy than both synchronous SGD and asynchronous SGD, and nearly approaches the performance of sequential SGD. As for the future work, we plan to test DC-ASGD on larger computer clusters, where with the increasing number of local workers, the delay will become more serious. Furthermore, we will investigate the economical approximation of higher-order items in the Taylor expansion to achieve more effective delay compensation.

	\small
	\bibliography{dcasgd}
	\bibliographystyle{icml2017}
	
	\normalsize
	\newpage
	\onecolumn
		
	\appendix
			
	\icmltitle{Supplementary Material: Asynchronous Stochastic Gradient Descent with Delay Compensation}
	
	\section{Theorem 3.1 and Its Proof}
	\textbf{Theorem 3.1:}
	
	\textit{Assume the loss function is $L_1$-Lipschitz. If {\small$\lambda\in [0, 1]$} make the following inequality holds, {\small\begin{equation}\label{eq99}\sum_{k=1}^{K}\frac{1}{\sigma_k^3(x,\textbf{w}_t)}\geq2\left[C_{ij}\left(\sum_{k=1}^{K}\frac{1}{\sigma_k(x,\textbf{w}_t)}\right)^2+C_{ij}^{'}L_1^2|\epsilon_t|\right],\end{equation}} where $C_{ij}=\frac{1}{1+\lambda}(\frac{u_iu_j\beta}{l_il_j\sqrt{\alpha}})^2$, $C_{ij}^{'}=\frac{1}{(1+\lambda)\alpha(l_il_j)^2}$, and the model converges to the optimal model, then the MSE of $\lambda G(\textbf{w}_t)$ is smaller than the MSE of $G(\textbf{w}_t)$ in approximating Hessian $H(\textbf{w}_t)$.}
	
	\textbf{Proof:}
	
	For simplicity, we abbreviate $\mathbb{E}_{(Y|x,w^*)}$ as $\mathbb{E}$, $G_t$ as $G(\textbf{w}_t)$ and $H_t$ as $H(\textbf{w}_t)$. 
	First, we calculate the {MSE} of $G_t$, $\lambda G_t$ to approximate $H_t$ for each element of $G_t$. We denote the element in the $i$-th row and $j$-th column of $G(w_t)$ as $G_{ij}^t$ and $H(w_t)$ as $H_{ij}(t)$.
	
	The MSE of $G_{ij}^t$:
	{\small\begin{align}
		\quad &\mathbb{E}(G_{ij}^t-\mathbb{E}H_{ij}^t)^2=\mathbb{E}(G_{ij}^t-\mathbb{E}G_{ij}^t)^2+(\mathbb{E}H_{ij}^t-\mathbb{E}G_{ij}^t)^2=\mathbb{E}(G_{ij}^t)^2-(\mathbb{E}G_{ij}^t)^2+\epsilon_t^2\label{eq221}
		\end{align}}
	
	The MSE of $\lambda g_{ij}$:
	{\small\begin{align}
		\quad \mathbb{E}(\lambda G_{ij}^t-\mathbb{E}H_{ij}^t)^2&=\lambda^2\mathbb{E}(G_{ij}^t-\mathbb{E}G_{ij}^t)^2+(\mathbb{E}H_{ij}^t-\lambda\mathbb{E}G_{ij}^t)^2 \nonumber\\
		&=\lambda^2\mathbb{E}(G_{ij}^t)^2-\lambda^2(\mathbb{E}G_{ij}^t)^2+(1-\lambda)^2(\mathbb{E}G_{ij}^t)^2+\epsilon_t^2+2(\lambda-1)\mathbb{E}G_{ij}^t\epsilon_t \label{eq23}
		\end{align}}
	The condition for $\mathbb{E}(G_{ij}^t-\mathbb{E}H_{ij}^t)^2\geq \mathbb{E}(\lambda G_{ij}^t-\mathbb{E}H_{ij}^t)^2$ is 
	{\small\begin{align}\label{eq(4)}
		(1-\lambda^2)(\mathbb{E}(G_{ij}^t)^2-(\mathbb{E}G_{ij}^t)^2)\geq 2(1-\lambda)(\mathbb{E}G_{ij}^t)^2+2(\lambda-1)\mathbb{E}G_{ij}^t\epsilon_t
		\end{align}} 
	
	Inequality (\ref{eq(4)}) is equivalent to 
	{\small\begin{align}\label{eq(5)}
		(1+\lambda)\mathbb{E}(G_{ij}^t)^2\geq 2[(\mathbb{E}G_{ij}^t)^2-\mathbb{E}G_{ij}^t\epsilon_t]
		\end{align}} 
	Next we calculate $\mathbb{E}(G_{ij}^t)^2$, and $(\mathbb{E}G_{ij}^t)^2$ which appear in Eqn.(\ref{eq(5)}). For simplicity, we denote $\sigma_k(x,\textbf{w}_t)$ as $\sigma_k$, and $I_{[Y=k]}$ as $z_k$. Then we can get:
	{\small\begin{align}
		\mathbb{E}(g_{ij})^2&=\mathbb{E}_{(Y|x,\textbf{w}_t)}\left(\frac{\partial}{\partial w_i}\log{P(Y|x,\textbf{w}_t)}\right)^2\left(\frac{\partial}{\partial w_j}\log{P(Y|x,\textbf{w}_t)}\right)^2\\
		&\geq\mathbb{E}_{(Y|x,\textbf{w}^*)}\left(\sum_{k=1}^{K}\left(-\frac{z_k}{\sigma_k}\right)\right)^4\left(l_il_j\right)^2\nonumber\\
		&=\alpha\left(l_il_j\right)^2\left(\sum_{k=1}^{K}\frac{1}{\sigma_k^3(x,\textbf{w}_t)}\right)\label{eq25}\\
		(\mathbb{E}h_{ij})^2
		&=\left(\mathbb{E}_{(Y|x,\textbf{w}^*)}\sum_{k=1}^{K}\frac{\partial \sigma_k}{\partial w_i}\left(-\frac{z_k}{\sigma_k}\right)\cdot\sum_{k=1}^{K}\frac{\partial \sigma_k}{\partial w_j}\left(-\frac{z_k}{\sigma_k}\right)\right)^2\nonumber\\
		&\leq \beta^2\left(u_iu_j\right)^2\left(\sum_{k=1}^{K}\frac{1}{\sigma_k(x,\textbf{w}_t)}\right)^2.\label{eq27}
		\end{align}}
	
	By substituting Ineq.(\ref{eq25}) and Ineq.(\ref{eq27}) into Ineq.(\ref{eq(5)}), a sufficient condition for Ineq.(\ref{eq(5)}) to be satisfied is  {\small$\sum_{k=1}^{K}\frac{1}{\sigma_k^3(x,\textbf{w}_t)}\geq2\left[C_{ij}\left(\sum_{k=1}^{K}\frac{1}{\sigma_k(x,\textbf{w}_t)}\right)^2+C_{ij}^{'}L_1^2|\epsilon_t|\right]$} because $G_{ij}^t\leq L_1^2$.
	\ \ $\Box$
	
	\section{Corollary 3.2 and Its Proof}
	\textbf{Corollary 3.2:}
	\textit{A sufficient condition for inequality (\ref{eq99}) is $\lambda\in[0,1]$ and $\exists k_0\in[K]$ such that {\small$\sigma_{k_0}\in\left[1-\frac{K-1}{2(C_{ij}K^2+C_{ij}^{'}L_1^2\epsilon_t)},1\right]$}.}
	
	\textbf{Proof:}\\
	Denote $\Delta=\frac{K-1}{2C_{ij}K^2}$ and $F(\sigma_1,...,\sigma_K)=\sum_{k=1}^{K}\frac{1}{\sigma_k^3(x,\textbf{w}_t)}-2C_{ij}\left(\sum_{k=1}^{K}\frac{1}{\sigma_k(x,\textbf{w}_t)}\right)^2-2C_{ij}^{'}L_1^2|\epsilon_t|$. 	If $\exists k_1\in[K]$ such that {\small$\sigma_{k_1}\in\left[1-\Delta,1\right]$}, we have for $k\neq k_1$ {\small$\sigma_{k}\in\left[0,\Delta\right]$}.
	Therefore 
	\begin{align}
	F(\sigma_1,...,\sigma_{K})&\geq\frac{1}{(\sigma_{k_1})^3}+\frac{K-1}{\Delta^3}-2C_{ij}\left(\frac{1}{\sigma_{k_1}}+\frac{K-1}{\Delta}\right)^2-2C_{ij}^{'}L_1^2|\epsilon_t|\\
	&\geq\frac{K-1}{\Delta^3}-2C_{ij}\left(\left(\frac{K-1}{\Delta}\right)^2+\frac{1}{\sigma^2_{k_1}}+\frac{2(K-1)}{\sigma_{k_1}\Delta}\right)-2C_{ij}^{'}L_1^2|\epsilon_t|\label{eq321}\\
	&\geq\frac{K-1}{\Delta^3}-2C_{ij}\left(\frac{(K-1)^2}{\Delta^2}+\frac{2K-1}{\sigma_{k_1}\Delta}\right)-2C_{ij}^{'}L_1^2|\epsilon_t|\label{eq32}\\
	&=\frac{1}{\Delta}\left(\frac{K-1}{\Delta^2}-2C_{ij}\left(\frac{(K-1)^2}{\Delta}+\frac{2K-1}{\sigma_{k_1}}\right)\right)-2C_{ij}^{'}L_1^2|\epsilon_t|\label{eq323}\\
	&\geq \frac{1}{\Delta}\left(\frac{K-1}{\Delta^2}-2C_{ij}\left(\frac{(K-1)^2+2K-1}{\Delta}\right)\right)-2C_{ij}^{'}L_1^2|\epsilon_t|\label{eq324}\\
	&\geq \frac{1}{\Delta^2}\left(\frac{K-1}{\Delta}-2C_{ij}K^2-2C_{ij}^{'}L_1^2|\epsilon_t|\right)\label{eq35}\\
	&=0 \label{eq36}\\\nonumber
	\end{align}
	where Ineq.(\ref{eq32}) and (\ref{eq324}) is established since $\sigma_{k_1}>\Delta$; and Eqn.(\ref{eq36}) is established by putting $\Delta=\frac{K-1}{2(C_{ij}K^2+C_{ij}^{'}L_1^2|\epsilon_t|)}$ in Eqn.(\ref{eq35}). \ \ $\Box$
	
	\section{Uniform upper bound of MSE}
	\begin{lemma}
		Assume the loss function is $L_1$-Lipschitz, and the diagonalization error of Hessian is upper bounded by $\epsilon_D$, i.e., $||Diag(H(\textbf{w}_t))- H(\textbf{w}_t)||\leq \epsilon_D$, \footnote{\cite{yann1987modeles} demonstrated that the diagonal approximation to Hessian for neural networks is an efficient method with no much drop on accuracy} then we have, for $\forall t$,  
		{\small\begin{align}
			mse^t(Diag(\lambda G))\leq 4\lambda^2 V_1+4(1-\lambda)^2L_1^4+4\epsilon_t^2+4\epsilon_D,
			\end{align}}
		where $V_1$ is the upper bound of the variance of $G(\textbf{w}_t)$.
	\end{lemma}
	\textbf{Proof:}
	{\small\begin{align}
		&mse^t(Diag(\lambda G))\\
		\leq&\mathbb{E}\|Diag(\lambda G(w_t))-H(w_t)\|^2\\
		\leq&4\mathbb{E}\|Diag(\lambda G(w_t))-\mathbb{E}(Diag(\lambda G(w_t)))\|^2+4\|\mathbb{E}(Diag(\lambda G(w_t)))-\mathbb{E}(Diag(G(w_t)))\|^2\\
		&+4\|\mathbb{E}(Diag(G(w_t)))-\mathbb{E}(Diag(H(w_t)))\|^2+4\|\mathbb{E}(Diag(H(w_t)))-\mathbb{E}H(w_t)\|^2\\
		\leq&4\lambda^2 V_1+4(1-\lambda)^2L_1^4+4\epsilon_t^2+4\epsilon_D
		\end{align}}
	
	\section{Convergence Rate for DC-ASGD: Convex Case}
	DC-ASGD is a general method to compensate delay in ASGD. We first show the convergence rate for convex loss function. If the loss function $f(w)$ is convex about $w$, we can add a regularization term $\frac{\rho}{2}\|w\|^2$ to make the objective function $F(w)+\frac{\rho}{2}\|w\|^2$ strongly convex. Thus, we assume that the objective function is $\mu$-strongly convex.
	
	\textbf{Theorem 4.1: (Strongly Convex)}
	\textit{If $f(w)$ is $L_2$-smooth and $\mu$-strongly convex about $w$, $\nabla f(w)$ is $L_3$-smooth about $w$ and the expectation of the $\|\cdot\|_2^2$ norm of the delay compensated gradient is upper bounded by a constant $G$. 
		By setting the learning rate $\eta_t=\frac{1}{\mu t}$, DC-ASGD has convergence rate as $$	\mathbb{E}F(w_{t})-F(w^*)\leq\frac{2L_2^2G^2}{t\mu^4}\left(1+4\tau C_{\lambda}\right)+\frac{2G^2L_2^2\theta\sqrt{\tau}}{\mu^4t\sqrt{t}}+\frac{L^3L_2^3\tau^2G^3}{\mu^6t^2},$$	where $\theta=\frac{2HKLG}{\mu}\sqrt{\frac{L_2}{\mu}(1+\frac{\tau GL_3}{\mu L_2})}$ and $C_{\lambda}=(1-\lambda)L_1^2+\epsilon_D$, and the expectation is taking with respect to the random sampling of DC-ASGD  and $\mathbb{E}_{(y|x,w^*)}$.}

	\textbf{Proof:}
	
	We denote $g^{dc}(w_t)=g(w_t)+\lambda g(w_t)\odot g(w_t)\odot(w_{t+\tau}-w_t)$, $g^{h}(w_t)=g(w_t)+H_{i_t}(w_t)(w_{t+\tau}-w_t)$ and $\nabla F^{h}(w_t)=\nabla F(w_t)+\mathbb{E}_{i_t}H_{i_t}(w_t)(w_{t+\tau}-w_t)$. Obviously, we have $\mathbb{E}g^{h}(w_t)=\nabla F^{h}(w_t)$.
	By the smoothness condition, we have
	{\small\begin{eqnarray}
		&&\mathbb{E}F(w_{t+\tau+1})-F(w^*)\\
		&\leq&F(w_{t+\tau})-F(w^*)-\langle\nabla F(w_{t+\tau}),w_{t+\tau+1}-w_{t+\tau}\rangle+\frac{L_2}{2}\|w_{t+\tau+1}-w_{t+\tau}\|^2\\
		&\leq&F(w_{t+\tau})-F(w^*)-\eta_{t+\tau}\langle\nabla F(w_{t+\tau}),g^{dc}(w_t)\rangle+\frac{L_2\eta_{t+\tau}^2G^2}{2}\label{eq41}\\
		&=&F(w_{t+\tau})-F(w^*)-\eta_{t+\tau}\langle\nabla F(w_{t+\tau}),\nabla F(w_{t+\tau})\rangle+\eta_{t+\tau}\langle\nabla F(w_{t+\tau}),\nabla F(w_{t+\tau})-\nabla F^{h}(w_{t})\rangle\\
		&&+\eta_{t+\tau}\langle\nabla F(w_{t+\tau}),\mathbb{E}g^{h}(w_{t})-g^{dc}(w_t)\rangle+\frac{L_2\eta_{t+\tau}^2G^2}{2} \label{eq20}
		\end{eqnarray}}
	Since $f(w)$ is $L_2$-smooth and $\mu$ strongly convex, we have 
	{\small\begin{align}
		-\langle\nabla F(w_{t+\tau}),\nabla F(w_{t+\tau})\rangle\leq -\mu^2\|w_{t+\tau}-w^*\|^2\leq-\frac{2\mu^2}{L_2}(F(w_{t+\tau})-F(w^*)).\label{eq22}
		\end{align}}
	
	For the term {\small$\eta_{t+\tau}\langle\nabla F(w_{t+\tau}),\nabla F(w_{t+\tau})-\nabla F^{h}(w_{t})\rangle$,} we have 
	{\small\begin{eqnarray}
		&&\eta_{t+\tau}\langle\nabla F(w_{t+\tau}),\nabla F(w_{t+\tau})-\nabla F^{h}(w_{t})\rangle\\
		&\leq&\eta_{t+\tau}\|\nabla F(w_{t+\tau})\|\|\nabla F(w_{t+\tau})-\nabla F^{h}(w_t)\|\\
		&\leq&\eta_{t+\tau}G\|\nabla F(w_{t+\tau})-\nabla F^{h}(w_t)\|
		\end{eqnarray}}
	By the smoothness condition for $\nabla F(w)$, we have
	{\small\begin{align}
		\|\nabla F(w_{t+\tau})-\nabla F^{h}(w_t)\|\leq \frac{L_3}{2}\|w_{t+\tau}-w_t\|^2\leq\frac{L_3\tau G^2}{2}\sum_{j=0}^{\tau-1}\eta_{t+j}^2 \label{eq26}
		\end{align}}
	Let $\eta_t=\frac{L_2}{\mu^2 t}$, we can get $\sum_{j=1}^{\tau}\eta_{t+j}^2\leq\frac{L_2^2}{\mu^4}\cdot\frac{\tau}{t(t+\tau)}\leq\frac{2L_2^2\tau}{\mu^4(t+\tau)^2}$. 
	
	For the term $\eta_{t+\tau}\langle\nabla F(w_{t+\tau}),\mathbb{E}g^{h}(w_{t})-g^{dc}(w_t)\rangle$, we have 	
	{\small\begin{align}
		&\langle\nabla F(w_{t+\tau}),\mathbb{E}(g^{h}(w_{t})-g^{dc}(w_t))\rangle\\
		&\leq\|\nabla F(w_{t+\tau})\|\|\mathbb{E}(\lambda g(w_t)\odot g(w_t)-H(w_t))(w_{t+\tau}-w_t)\|\\
		&\leq G^2\tau\sum_{j=0}^{\tau-1}\eta_{t+j}(\|\mathbb{E}(\lambda g(w_t)\odot g(w_t)-g(w_t)\odot g(w_t) \|+\|g(w_t)\odot g(w_t)-Diag (H(w_t))\|+\|Diag (H(w_t))-H(w_t)\|)\\
		&\leq \frac{2G^2L_2\tau}{(t+\tau)\mu^2}(C_{\lambda}+\epsilon_t),
		\end{align}}
	where $C_{\lambda}=(1-\lambda)L_1^2+\epsilon_D$.
	
	Using Lemma \ref{lemma1}, $\epsilon_t\leq \theta\sqrt{\frac{1}{t}}\leq\theta\sqrt{\frac{\tau}{t+\tau}}$.
	Putting inequality \ref{eq22} and \ref{eq26} in inequality \ref{eq20}, we have
	{\small\begin{align}
		\mathbb{E}F(w_{t+\tau+1})-F(w^*)&\leq\left(1-\frac{2}{t+\tau}\right)(\mathbb{E}F(w_{t})-F(w^*))+\frac{L_3L_2^3\tau^2 G^3}{\mu^6(t+\tau)^3}\\
		&+\frac{2G^2L_2^2\tau}{\mu^4(t+\tau)^2}\left(C_{\lambda}+\theta\sqrt{\frac{\tau}{t+\tau}}\right)+\frac{L_2^2G^2}{2(t+\tau)^2\mu^4}
		\end{align}}
	We can get 
	{\small\begin{eqnarray}
		\mathbb{E}F(w_{t})-F(w^*)\leq\frac{2L_2^2G^2}{t\mu^4}\left(1+4\tau C_{\lambda}\right)+\frac{2G^2L_2^2\theta\sqrt{\tau}}{\mu^4t\sqrt{t}}+\frac{L^3L_2^3\tau^2G^3}{\mu^6t^2}.
		\end{eqnarray}}
	by induction. $\Box$

	\textbf{Discussion:} 
	
	(1). Following the above proof steps and using $\|\nabla F(w_{t+\tau})-\nabla F(w_t)\|\leq L_2\|w_{t+\tau}-w_t\|$, we can get the convergence rate of ASGD is 
	{\small\begin{align}
		\mathbb{E}F(w_{t})-F(w^*)\leq\frac{2L_2^2G^2}{t\mu^4}\left(1+4\tau L_2\right).
		\end{align}} Compared the convergence rate of DC-ASGD with ASGD, the extra term $\frac{2G^2L_2^2\theta\sqrt{\tau}}{\mu^4t\sqrt{t}}+\frac{L^3L_2^3\tau^2G^3}{\mu^6t^2}$ converge to zero faster than $\frac{2L_2^2G^2}{t\mu^4}\left(1+4\tau C_{\lambda}\right)$ in terms of the order of $t$. Thus, when $t$ is large, the extra term has smaller value. We assume that $t$ is large and the term can be neglected. Then the condition for DC-ASGD outperforming ASGD is $L_2>C_{\lambda}$.

	\section{Convergence Rate for DC-ASGD: Nonconvex Case}

	\textbf{Theorem 5.1: (Nonconvex Case)}
	\textit{Assume that Assumptions 1-4 hold. Set the learning rate
		{\small\begin{equation}\label{eq1}
			\eta_t=\sqrt{\frac{2(F(w_1)-F(w^*)}{bTV^2L_2}},
			\end{equation}} where $b$ is the mini-batch size, and $V$ is the upper bound of the variance of the delay-compensated gradient. If $T\geq \max\{\mathcal{O}(1/r^{4}), 2D_0bL_2/V^2\}$ and delay $\tau$ is upper-bounded as below,
		{\small\begin{align}
			&\tau\leq\min\left\{\frac{L_2V}{C_{\lambda}}\sqrt{\frac{L_2 T}{2D_0b}},\frac{V}{C_{\lambda}}\sqrt{\frac{L_2 T}{2D_0b}},\frac{TV}{\tilde{C}}\sqrt{\frac{L_2}{bD_0}},\frac{VL_2T}{4\tilde{C}}\sqrt{\frac{TL_2}{2D_0b}}\right\}.
			\end{align}}  then DC-ASGD has the following ergodic convergence rate,
		\begin{equation}
		\min_{t=\{1,\cdots,T\}}\mathbb{E}(\|\nabla F(\textbf{w}_t)\|^2)\leq V\sqrt{\frac{2D_0L_2}{bT}},
		\end{equation}
		where the expectation is taken with respect to the random sampling in  SGD and the data distribution $P(Y|x,\textbf{w}^*)$.}
	
	\textbf{Proof:}
	
	We denote $g_m(w_{t})+\lambda g_m(w_t)\odot g_m(w_t)\odot(w_{t+\tau}-w_t)$ as $g_m^{dc}(w_{t})$ where $m\in\{1,\cdots,b\}$ is the index of instances in the minibatch. From the proof the Theorem 1 in ASGD \cite{lian2015asynchronous}, we can get 
	{\small\begin{eqnarray}
		&&\mathbb{E}F(w_{t+\tau+1})-F(w_{t+\tau})\\
		&\leq&\langle\nabla F(w_{t+\tau}), w_{t+\tau}-w_t\rangle+\frac{L_2}{2}\|w_{t+\tau+1}-w_{t+\tau}\|^2\\
		&\leq&-\eta_{t+\tau}\langle\nabla F(w_{t+\tau}),\sum_{m=1}^b\mathbb{E}g_m^{dc}(w_t)\rangle+\frac{\eta_{t+\tau}^2L_2}{2}\mathbb{E}\left(\left\|\sum_{m=1}^bg_m^{dc}(w_t)\right\|^2\right)\\
		&\leq&-\frac{b\eta_{t+\tau}}{2}\left(\|\nabla F(w_{t+\tau})\|^2+\left\|\sum_{m=1}^b\mathbb{E}g_m^{dc}(w_t)\right\|^2-\left\|\nabla F(w_{t+\tau})-\sum_{m=1}^b\mathbb{E}g_m^{dc}(w_t)\right\|^2\right) \nonumber\\
		&&+\frac{\eta_{t+\tau}^2L_2}{2}\mathbb{E}\left(\left\|\sum_{m=1}^bg_m^{dc}(w_t)\right\|^2\right)\label{eq60}
		\end{eqnarray}}
	For the term $T_1=\left\|\nabla F(w_{t+\tau})-\sum_{m=1}^b\mathbb{E}g_m^{dc}(w_t)\right\|^2$, by using the smooth condition of $g$, we have 
	{\small\begin{eqnarray}
		T_1&=&\left\|\nabla F(w_{t+\tau})-\sum_{m=1}^b\mathbb{E}g_m^{dc}(w_t)\right\|^2\\
		&\leq&\left\|\nabla F(w_{t+\tau})-\nabla F^{h}(w_t)+\nabla F^{h}(w_t)-\sum_{m=1}^b\mathbb{E}g_m^{dc}(w_t)\right\|^2 \\
		&\leq&2\left\|\frac{L_3}{2}\|w_{t+\tau}-w_t\|^2\right\|^2+2\left\|\nabla F^{h}(w_t)-\sum_{m=1}^b\mathbb{E}g_m^{dc}(w_t)\right\|^2\\
		&\leq&(L_3^2\pi^2/2+2(((1-\lambda)L_1^2+\epsilon_D)^2+\epsilon_t^2))\|w_{t+\tau}-w_t\|^2
		\end{eqnarray}}
	Thus by following the proof of ASGD, we have 
	{\small\begin{eqnarray}
		\mathbb{E}(T_1)\leq 4(L_3^2\pi^2/4+((1-\lambda)L_1^2+\epsilon_D)^2+\epsilon_t^2)\left(b\tau\eta_{t+\tau}^2V^2+\tau^2\eta_{t+\tau}^2\left\|b\mathbb{E}g_m^{dc}(w_t)\right\|^2\right).\label{eq3}
		\end{eqnarray}}
	For the term $T_2=\mathbb{E}\left(\left\|\sum_{m=1}^bg_m^{dc}(w_t)\right\|^2\right)$, it has
	{\small\begin{eqnarray}
		\mathbb{E}(T_2)\leq bV^2+\left\|b\mathbb{E}g_m^{dc}(w_t)\right\|^2.\label{eq4}
		\end{eqnarray}}
	By putting Ineq.(\ref{eq3}) and Ineq.(\ref{eq4}) in Ineq.(\ref{eq60}), we can get
	{\small\begin{eqnarray}
		&&\mathbb{E}(F(w_{t+\tau+1})-F(w_{t+\tau})\\
		\leq&&-\frac{b\eta_{t+\tau}}{2}\mathbb{E}\|\nabla F(w_{t+\tau})\|^2+\left(\frac{\eta_{t+\tau}^2L_2}{2}-\frac{\eta_{t+\tau}}{2b}\right)\mathbb{E}\left(\left\|b\mathbb{E}g_m^{dc}(w_t)\right\|^2\right) \nonumber\\
		&&+\left(\frac{\eta_{t+\tau}^2bL_2}{2}+(L_3^2\pi^2/2+2((1-\lambda)L_1^2+\epsilon_D)^2+\epsilon_t^2)b^2\tau\eta_{t+\tau}^3\right)V^2\\
		&&+(L_3^2\pi^2/2+2((1-\lambda)L_1^2+\epsilon_D)^2+\epsilon_t^2)b\tau^2\eta_{t+\tau}^3\mathbb{E}\left(\left\|b\mathbb{E}g_m^{dc}(w_t)\right\|^2\right)\label{eq5}
		\end{eqnarray}} 
	Summarizing the Ineq.(\ref{eq5}) from $t=1$ to $t+\tau=T$, we have
	{\small\begin{align}
		&\mathbb{E}F(w_{T+1})-F(w_1)\\
		\leq&-\frac{b}{2}\sum_{t=1}^T\eta_t\mathbb{E}\|\nabla F(w_t)\|^2+\sum_{t=1}^T\left(\frac{\eta_{t+\tau}^2bL_2}{2}+(L_3^2\pi^2/2+2((1-\lambda)L_1^2+\epsilon_D)^2+\epsilon_t^2)b^2\tau\eta_{t+\tau}^3\right)V^2\\
		&+\sum_{t=1}^T\left(\frac{\eta_t^2L_2}{2}+(L_3^2\pi^2/2+2((1-\lambda)L_1^2+\epsilon_D)^2+\epsilon_t^2)b\tau^2\eta_t^3-\frac{\eta_t}{2b}\right)\mathbb{E}\left\|b\mathbb{E}g_m^{dc}(w_{\max\{t-\tau,1\}})\right\|^2.
		\end{align}}By Lemma \ref{lemma1} and under our assumptions, we have when $t>T_0$, $w_t$ will goes into a strongly convex neighbourhood of some local optimal $w_{loc}$. Thus, $\epsilon_t\leq\epsilon_{nc}+\theta\sqrt{1/(t-T_0)}$, when $t>T_0$ and $\epsilon_t<\max_{s\in{1,\cdots, T_0}}\epsilon_s$ when $t<T_0$.
	
	Let $\eta_t=\sqrt{\frac{2(F(w_1)-F(w^*)}{bTV^2L_2}}$. It follows that 
	{\small\begin{align}
		&\sum_{t=1}^T\frac{\eta_t L_2}{2}+(L_3^2\pi^2/2+2((1-\lambda)L_1^2+\epsilon_D)^2+\epsilon_t^2)b\tau^2\eta_t^2\\
		\leq&\sum_{t=1}^T\left\{\frac{\eta_t L_2}{2}+(L_3^2\pi^2/2+2((1-\lambda)L_1^2+\epsilon_D)^2+2\epsilon_{nc}^2)b\tau^2\eta_t^2\right\}+2b\tau^2\eta_t^2(4T_0\max_{s\in{1,\cdots, T_0}}(\epsilon_s)^2+4\theta^2\log(T-T_0))
		\end{align}}
	
	We ignore the $\log(T-T_0)$ term and regards $\tilde{C}^2=4T_0\max_{s\in{1,\cdots, T_0}}(\epsilon_s)^2+4\theta^2\log(T-T_0)$ as a constant, which yields
	{\small\begin{align}
		&\sum_{t=1}^T\frac{\eta_t L_2}{2}+(L_3^2\pi^2/2+2((1-\lambda)L_1^2+\epsilon_D)^2+\epsilon_t^2)b\tau^2\eta_t^2\\
		\leq&\sum_{t=1}^T\left\{\frac{\eta_t L_2}{2}+(L_3^2\pi^2/2+2((1-\lambda)L_1^2+\epsilon_D)^2+2\epsilon_{nc}^2)b\tau^2\eta_t^2\right\}+2\tau^2\eta_{t}^2b\tilde{C}^2
		\end{align}}
	
	$\eta_t$ should be set to make \begin{align}\label{eq67}\sum_{t=1}^T\left(\frac{\eta_t^2L_2}{2}+(L_3^2\pi^2/2+2((1-\lambda)L_1^2+\epsilon_D)^2+2\epsilon_{nc}^2)b\tau^2\eta_t^3+\frac{2\tau^2\eta_t^3b\tilde{C}^2}{T}-\frac{\eta_t}{2b}\right)\leq 0.\end{align}
	
	Then we can get 
	{\small\begin{align}
		&\frac{1}{T}\sum_{t=1}^{T}\mathbb{E}\|\nabla F(w_t)\|^2\\
		\leq&\frac{2(F(w_1)-F(w^*)+Tb(\eta_t^2 L_2+2(L_3^2\pi^2/2+2((1-\lambda)L_1^2+\epsilon_D)^2+2\epsilon_{nc}^2)b\tau\eta_t^3)V^2+\frac{\eta_t^3\tilde{C}^24b\tau}{T}V^2}{bT\eta_t}\\
		\leq&\frac{2(F(w_1)-F(w^*)}{bT\eta_t}+(\eta_t L_2+2(L_3^2\pi^2/2+2((1-\lambda)L_1^2+\epsilon_D)^2+2\epsilon_{nc}^2)b\tau\eta_t^2)V^2+\frac{\eta_t^2\tilde{C}^24b\tau V^2}{T}\\
		\end{align}}
	We set $\eta_t$ to make 
	{\small\begin{align}
		(2(L_3^2\pi^2/2+2((1-\lambda)L_1^2+\epsilon_D)^2+2\epsilon_{nc}^2)b\tau\eta_t^2)+\frac{\eta_t^2\tilde{C}^24b\tau }{T}\leq\eta_tL_2 \label{eq66}
		\end{align}}
	
	Thus let $\eta_t=\sqrt{\frac{2(F(w_1)-F(w^*)}{bTV^2L_2}}$, 	{\small\begin{align}\label{eq68}
		&\frac{1}{T}\sum_{t=1}^{T}\mathbb{E}\|\nabla F(w_t)\|^2
		\leq V\sqrt{\frac{2D_0L_2}{bT}}.
		\end{align}}
	And we can get the condition for $T$ by putting $\eta$ in ineq.\ref{eq67} and ineq.\ref{eq66}, we can get that {\small\begin{align}
		&\tau\leq\min\left\{\frac{L_2V}{C_{\lambda}}\sqrt{\frac{L_2 T}{2D_0b}},\frac{V}{C_{\lambda}}\sqrt{\frac{L_2 T}{2D_0b}},\frac{TV}{\tilde{C}}\sqrt{\frac{L_2}{bD_0}},\frac{VL_2T}{4\tilde{C}}\sqrt{\frac{TL_2}{2D_0b}}\right\}.
		\end{align}}

	\section{Decreasing rate of the approximation error $\epsilon_t$}
	Since $\epsilon_t$ is contained the proof of the convergence rate for DC-ASGD , in this section we will introduce a lemma which describes the approximation error $\epsilon_t$ the for both convex and nonconvex cases.
	\begin{lemma}\label{lemma1}
		Assume that the true label $y$ is generated according to the distribution {\small$\mathbb{P}(Y=k|x,w^*)=\sigma_k(x,w^*)$} and {\small$f(x,y,\textbf{w})=-\sum_{k=1}^K(I_{[y=k]}\log{\sigma_k(x;\textbf{w})})$}. 
		If we assume that the loss function is $\mu$-strongly convex about $w$. We denote $\textbf{w}_t$ is the output of DC-ASGD by using the outerproduct approximation of Hessian, we have 
		{\small\begin{eqnarray*}
				&&\epsilon_t=\Big|\mathbb{E}_{(x,y|\textbf{w}^*)}\frac{\partial^2}{\partial \textbf{w}^2}f(x,y,\textbf{w}_t)-\mathbb{E}_{(x,y|\textbf{w}^*)}\left(\frac{\partial}{\partial \textbf{w}}f(x,y,\textbf{w}_t)\right)\otimes\left(\frac{\partial}{\partial \textbf{w}}f(x,y,\textbf{w}_t)\right)\Big|\leq\theta\sqrt{\frac{1}{t}},
			\end{eqnarray*}}where $\theta=\frac{2HKLVL_2}{\mu^2}\sqrt{\frac{1}{\mu}(1+\frac{L_2+\lambda L_1^2}{L_2}\tau)}$. 
			
			If we assume that the loss function is $\mu$-strongly convex in a neighborhood of each local optimal $d(\textbf{w}_{loc}, r)$, $\left|\frac{\partial^2 \mathbb{P}(Y=k|x,\textbf{w})}{\partial^2\textbf{w}}\times\frac{1}{P(Y=k|x,w)}\right|\leq H$, $\forall k,x,w$, each $\sigma_k(\textbf{w})$ is $L$-Lipschitz continuous about $\textbf{w}$. We denote $\textbf{w}_t$ is the output of DC-ASGD by using the outerproduct approximation of Hessian, we have 
			{\small\begin{eqnarray*}
					&&\epsilon_t=\Big|\mathbb{E}_{(x,y|\textbf{w}^*)}\frac{\partial^2}{\partial \textbf{w}^2}f(x,y,\textbf{w}_t)-\mathbb{E}_{(x,y|\textbf{w}^*)}\left(\frac{\partial}{\partial \textbf{w}}f(x,y,\textbf{w}_t)\right)\otimes\left(\frac{\partial}{\partial \textbf{w}}f(x,y,\textbf{w}_t)\right)\Big|\leq\theta\sqrt{\frac{1}{t-T_0}}+\epsilon_{nc}.
				\end{eqnarray*}}
				where $t>T_0\geq \mathcal{O}(\frac{1}{r^8})$.
			\end{lemma}	
			\textbf{Proof:}
			{\small\begin{align}
				\mathbb{E}_{(y|x,\textbf{w}^*)}\frac{\partial^2}{\partial \textbf{w}^2}f(x,Y,\textbf{w}_t)&=-\mathbb{E}_{(y|x,\textbf{w}^*)}\frac{\partial^2}{\partial \textbf{w}^2}\left(\sum_{k=1}^K(I_{[y=k]}\log{\sigma_k(x;\textbf{w}_t)})\right) \nonumber\\
				&=-\mathbb{E}_{(y|x,\textbf{w}^*)}\frac{\partial^2}{\partial \textbf{w}^2}\log{\left(\prod_{k=1}^K\sigma_k(x,\textbf{w}_t)^{I_{[y=k]}}\right)} \nonumber\\
				&=-\mathbb{E}_{(y|x,\textbf{w}^*)}\frac{\partial^2}{\partial \textbf{w}^2}\log{\mathbb{P}(y|x,\textbf{w}_t)} \nonumber\\
				&=-\mathbb{E}_{(y|x,\textbf{w}^*)}\frac{\frac{\partial^2}{\partial\omega^2}\mathbb{P}(y|x,\textbf{w}_t)}{\mathbb{P}(y|x,\textbf{w}_t)}+\mathbb{E}_{(y|x,\textbf{w}^*)}\left(\frac{\frac{\partial}{\partial\omega}\mathbb{P}(y|x,\textbf{w}_t)}{\mathbb{P}(y|x,\textbf{w}_t)}\right)^2 \nonumber\\
				&=-\mathbb{E}_{(y|x,\textbf{w}^*)}\frac{\frac{\partial^2}{\partial\omega^2}\mathbb{P}(y|x,\textbf{w}_t)}{\mathbb{P}(y|x,\textbf{w}_t)}+\mathbb{E}_{(y|x,\textbf{w}^*)}\left(\frac{\partial}{\partial\omega}\log \mathbb{P}(y|x,\textbf{w}_t)\right)^2. \nonumber \\
				&=-\mathbb{E}_{(y|x,\textbf{w}^*)}\frac{\frac{\partial^2}{\partial\omega^2}\mathbb{P}(y|x,\textbf{w}_t)}{\mathbb{P}(y|x,\textbf{w}_t)}+\mathbb{E}_{(y|x,\textbf{w}^*)}\left(\frac{\partial}{\partial\omega}f(x,Y,\textbf{w}_t)\right)^2.\label{eq11}
				\end{align}}
			Since $\mathbb{E}_{(y|x,\textbf{w}_t)}\frac{\frac{\partial^2}{\partial\omega^2}\mathbb{P}(y|x,\textbf{w}_t)}{\mathbb{P}(y|x,\textbf{w}_t)}=0$ by the two equivalent methods to calculating fisher information matrix \cite{friedman2001elements},
			we have 
			{\small\begin{align}
				\left|\mathbb{E}_{(y|x,\textbf{w}^*)}\frac{\frac{\partial^2}{\partial\omega^2}\mathbb{P}(y|x,\textbf{w}_t)}{\mathbb{P}(y|x,\textbf{w}_t)}\right|
				&=\left|\mathbb{E}_{(y|x,\textbf{w}^*)}\frac{\frac{\partial^2}{\partial\omega^2}\mathbb{P}(y|x,\textbf{w}_t)}{\mathbb{P}(y|x,\textbf{w}_t)}-\mathbb{E}_{(y|x,\textbf{w}_t)}\frac{\frac{\partial^2}{\partial\omega^2}\mathbb{P}(y|x,\textbf{w}_t)}{\mathbb{P}(y|x,\textbf{w}_t)}\right| \nonumber\\
				&=\left|\sum_{k=1}^K\frac{\partial^2}{\partial\omega^2}\mathbb{P}(Y=k|X=x,\textbf{w}_t) \times \frac{\mathbb{P}(Y=k|x,\textbf{w}^*)-\mathbb{P}(Y=k|x,\textbf{w}_t)}{\mathbb{P}(Y=k|x,\textbf{w}_t)}\right|\label{eq10}\\\nonumber
				&\leq H\cdot\sum_{k=1}^K\left|\mathbb{P}(Y=k|x,\textbf{w}^*)-\mathbb{P}(Y=k|x,\textbf{w}_t)\right|\\
				&\leq HKL\|\textbf{w}_t-\textbf{w}_{loc}\|+HK\max_{k=1,\cdots,K}\left|\mathbb{P}(Y=k|x,\textbf{w}_{loc})-\mathbb{P}(Y=k|x,\textbf{w}^*)\right| \label{eq34}\\
				&\leq HKL\|\textbf{w}_t-\textbf{w}_{loc}\|+\epsilon_{nc}.
				\end{align}}

			For strongly convex objective functions, $\epsilon_{nc}=0$ and $w_{loc}=w^*$. The only thing we need is to prove the convergence of DC-ASGD without using the information of $\epsilon_t$ like before.
			By the smoothness condition, we have
			{\small\begin{eqnarray}
				&&\mathbb{E}F(w_{t+\tau+1})-F(w^*)\\
				&\leq&F(w_{t+\tau})-F(w^*)-\eta_{t+\tau}\langle\nabla F(w_{t+\tau}),\mathbb{E}g^{dc}(w_t)\rangle+\frac{L_2\eta_{t+\tau}^2V^2}{2}\label{eq31}\\
				&=&F(w_{t+\tau})-F(w^*)-\eta_{t+\tau}\langle\nabla F(w_{t+\tau}),\nabla F(w_{t+\tau})\rangle\\
				&&+\eta_{t+\tau}\langle\nabla F(w_{t+\tau}),\nabla F(w_{t+\tau})-\mathbb{E}g^{dc}(w_t)\rangle+\frac{L_2\eta_{t+\tau}^2V^2}{2}\\
				&\leq&(1-\frac{2\eta_{t+\tau}\mu^2}{L_2})(F(w_{t+\tau})-F(w^*))+\eta_{t+\tau}\|\nabla F(w_{t+\tau})\|\|\nabla F(w_{t+\tau})-\mathbb{E}g^{dc}(w_t)\|+\frac{L_2\eta_{t+\tau}^2V^2}{2}\\
				&\leq&(1-\frac{2\eta_{t+\tau}\mu^2}{L_2})(F(w_{t+\tau})-F(w^*))+\eta_{t+\tau}V\cdot(L_2+\lambda L_1^2)\|w_{t+\tau}-w_t\|+\frac{L_2\eta_{t+\tau}^2V^2}{2}\\
				&\leq&(1-\frac{2\eta_{t+\tau}\mu^2}{L_2})(F(w_{t+\tau})-F(w^*))+\eta_{t+\tau}V\cdot(L_2+\lambda L_1^2)\|\sum_{j=1}^\tau \eta_{t+\tau-j}g^{dc}(w_t)\|+\frac{L_2\eta_{t+\tau}^2V^2}{2}
				\end{eqnarray}}
			
			Taking expectation to the above inequality, we can get
			{\small\begin{eqnarray}
				\mathbb{E}F(w_{t+\tau+1})-F(w^*)&\leq&(1-\frac{2\eta_{t+\tau}\mu^2}{L_2})(\mathbb{E}F(w_{t+\tau})-F(w^*))+\frac{\eta_{t+\tau}^2(L_2+\lambda L_1^2)V^2\tau}{2}+\frac{L_2\eta_{t+\tau}^2V^2}{2}\\
				&\leq&(1-\frac{2\eta_{t+\tau}\mu^2}{L_2})(\mathbb{E}F(w_{t+\tau})-F(w^*))+\frac{\eta_{t+\tau}^2V^2L_2}{2}(1+\frac{L_2+\lambda L_1^2}{L_2}\tau).
				\end{eqnarray}}
			Let $\eta_t=\frac{L_2}{\mu^2 t}$, we have
			{\small\begin{eqnarray}
				\mathbb{E}F(w_{t+1})-F(w^*)\leq\left(1-\frac{2}{t}\right)(\mathbb{E}F(w_{t})-F(w^*))+\frac{V^2L_2^2}{2\mu^4t^2}\left(1+\frac{L_2+\lambda L_1^2}{L_2}\tau\right).
				\end{eqnarray}}
			We can get 
			{\small\begin{eqnarray}
				\mathbb{E}F(w_{t})-F(w^*)\leq\frac{2L_2^2V^2}{t\mu^4}\left(1+\frac{L_2+\lambda L_1^2}{L_2}\tau\right).
				\end{eqnarray}}
			by induction.
			Then we can get 
			{\small\begin{equation}\label{eq33}
				\|w_t-w^*\|^2\leq \frac{4L_2^2V^2}{t\mu^5}\left(1+\frac{L_2+\lambda L_1^2}{L_2}\tau\right).
				\end{equation}}
			By putting Ineq.{\ref{eq33}} into Ineq.{\ref{eq34}}, we can get the result in the theorem.

			For nonconvex case,	if $\textbf{w}_t\in\mathcal{B}(\textbf{w}_{loc},r)$, we have {\small$\mathbb{E}(\textbf{w}_t-\textbf{w}_{loc})\leq\frac{1}{\mu}\mathbb{E}\nabla F(\textbf{w}_t)$} under the assumptions. Next we will prove that, for nonconvex loss function $f(x,y,\textbf{w}_t)$, DC-ASGD has ergodic convergence rate. $\min_{t=1,\cdots,T}\mathbb{E}\|\frac{\partial}{\partial\textbf{w}_t}F(x,y,\textbf{w}_t)\|^2=\mathcal{O}(1/\sqrt{T})$, where the expectation is taking with respect to the stochastic sampling.

			Compared with the proof of ASGD \cite{lian2015asynchronous}, DC-ASGD with Hessian approximation has 
			{\small\begin{eqnarray}
				T_1&=&\|\nabla F(w_{t+\tau})-\mathbb{E}g^{dc}(w_t)\|^2\\
				&=&\|\nabla F(w_{t+\tau})-\nabla F(w_t)-\lambda \mathbb{E}g(w_t)\odot g(w_t)\cdot(w_{t+\tau}-w_t)\|^2\\
				&\leq&2\|\nabla F(w_{t+\tau})-\nabla F(w_t)\|^2+2\|\lambda \mathbb{E}g(w_t)\odot g(w_t)\cdot(w_{t+\tau}-w_t)\|^2\\
				&\leq&2(L_2^2+\lambda^2L_1^4)\|w_{t+\tau}-w_t\|^2 ,
				\end{eqnarray}}
			since $L_1$ is the upper bound of $\nabla f(w)$ and $L_2$ is the smooth coefficient of $f(w)$. Suppose that $\eta=\sqrt{\frac{2D_0}{bTV^2L_2}}$ and $\tau$ is upper bounded as Theorem 5.1, 
			{\small\begin{eqnarray}
				\min_{t=1,\cdots,T}\mathbb{E}\|\nabla F(w_t)\|^2\leq\frac{1}{T}\sum_{t=1}^{T}\mathbb{E}\|\nabla F(w_t)\|^2\leq \mathcal{O}(\frac{1}{T^{1/2}}).
				\end{eqnarray}}
			
			Referring to a recent work of Lee $et.al$ \cite{lee2016gradient}, GD with a random initialization and sufficiently small constant step size converges to a local minimizer almost surely under the assumptions in Theorem 1.2. Thus, the assumption that $F(w)$ is $\mu$-strongly convex in the $r$-neighborhood of arbitrary local minimum $w_{loc}$ is easily to be satistied with probability one. By the $L_1$-Lipschitz assumption, we have $P(Y=k|x,w_t)-P(Y=k|x,w_{loc})\leq L_1\|w_t-w_{loc}\|$. By the $L_2$-smooth assumption, we have $L_2\|w_t-w_{loc}\|^2\geq\langle\nabla F(w_t),w_t-w_{loc}\rangle$. Thus for $w_t\in\mathcal{B}(w_{loc},r)$, we have $\|\nabla F(w_t)\|\leq L_2\|w_t-w_{loc}\|\leq L_2r$. By the continuously twice differential assumption, we can assume that $\|\nabla F(w_t)\|\leq L_2\|w_t-w_{loc}\|\leq L_2 r$ for $w_t\in\mathcal{B}(w_{loc},r)$ and $\|\nabla F(w_t)\|\leq L_2\|w_t-w_{loc}\|> L_2r$ for $w_t\notin\mathcal{B}(w_{loc},r)$ without loss of generality \footnote{We can choose $r$ small enough to make it satisfied.}. Therefore $\min_{t=1,\cdots,T}\mathbb{E}\|\nabla F(w_t)\|^2\leq L_2^2r^2$ is a sufficient condition for $\mathbb{E}\|w_T-w_{loc}\|\leq r$. 
			{\small\begin{eqnarray}
				\min_{t=1,\cdots,T_0}\mathbb{E}\|\nabla F(w_t)\|^2\leq\mathcal{O}(\frac{1}{T_0^{1/2}}) \leq r^2.
				\end{eqnarray}}
			We have $T_0\geq \mathcal{O}\left(\frac{1}{r^4}\right)$.
			
			Thus we have finished the proof for nonconvex case.
			
			\begin{figure*}
				\centering
				\includegraphics[scale=0.34]{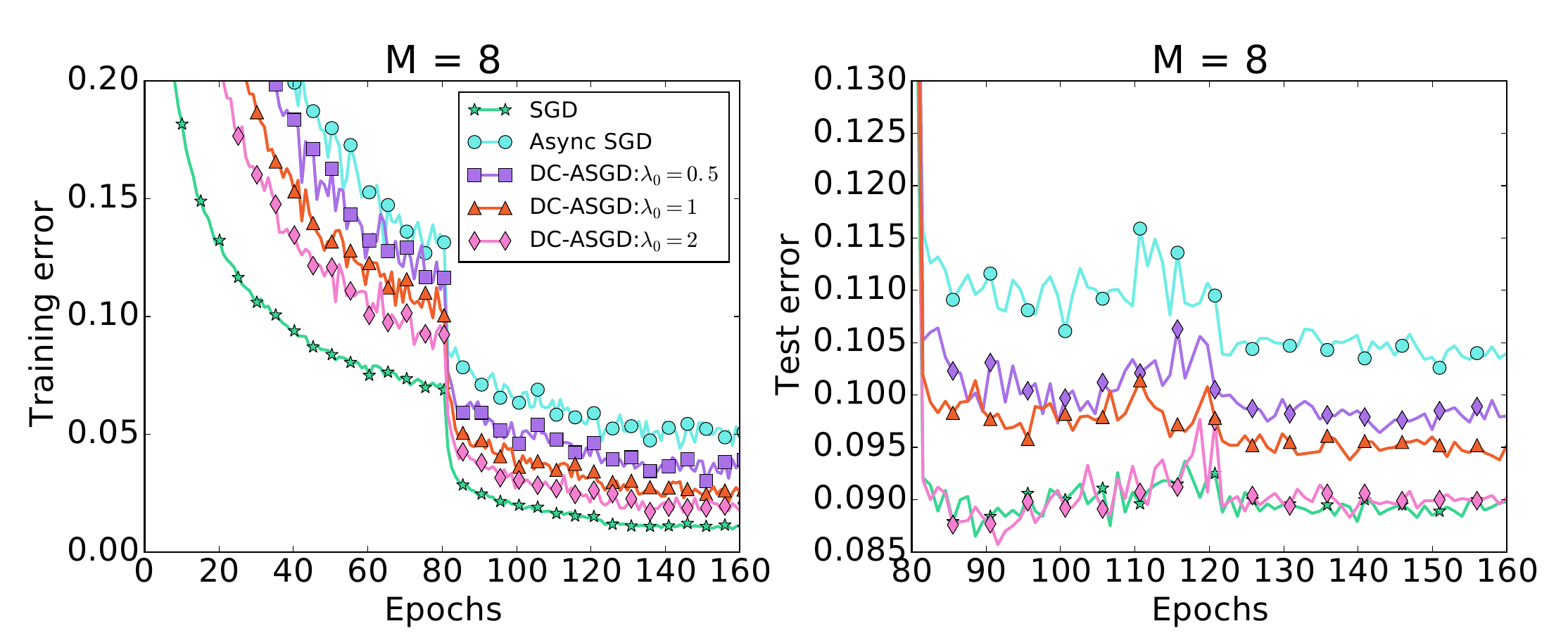}
				\caption{Error rates of the global model with Different $\lambda_0$ w.r.t. number of effective passes on CIFAR-10}
				\label{fig:lambda}
			\end{figure*}
			
			\section{Experimental Results on the Influence of $\lambda$}
			In this section, we show how the parameter $\lambda$ affect our DC-ASGD algorithm. We compare the performance of respectively sequential SGD, ASGD and DC-ASGD-a with different value of initial $\lambda_0$\footnote{We also compare different $\lambda_0$ for DC-ASGD-c and the results are very similar to DC-ASGD-a.}. The results are given in Figure \ref{fig:lambda}. This experiment reflects to the discussion in Section 5, too large value of this parameter ($\lambda_0 > 2$ in this setting) will introduce large variance and lead to a wrong gradient direction, meanwhile too small will make the compensation influence nearly disappear. As $\lambda$ decreasing, DC-ASGD will gradually degrade to ASGD. A proper $\lambda$ will lead to significant better accuracy. 
			
			\section{Large Mini-batch Synchronous SGD with Delay-Compensated Gradient}
			In this section, we discuss how delay-compensated gradient can be used in synchronous SGD. The effective mini-batch size in SSGD is usually enlarged $M$ times comparing with sequential SGD. A learning rate scaling trick is commonly used to overcome the influence of large mini-batch size in SSGD \cite{fbimagenet}: when the mini-batch size is multiplied by $M$, multiply the learning rate by $M$. For sequential mini-batch SGD with learning rate $\eta$ we have:
			{\begin{eqnarray}
				\textbf{w}_{t+M}=\textbf{w}_t-\eta \sum_{j=0}^{M-1} g(\textbf{w}_{t+j},z_{t+j}),\label{eq107}
				\end{eqnarray}}
			where $z_{t+j}$ is the $t+j$-th minibatch.
			
			On the other hand, taking one step with $M$ times large mini-batch size and learning rate $\hat{\eta} = M\eta$ in synchronous SGD yields:
			{\begin{eqnarray}
				\hat{\textbf{w}}_{t+1}=\textbf{w}_t-\hat{\eta} \frac{1}{M} \sum_{j=0}^{M-1} g(\textbf{w}_{t},z_t^j),\label{eq108}
				\end{eqnarray}} 
			where $z_t^j$ is the $t$-th minibatch on local machine $j$.
			
			Assume that $z_{t+j}=z_t^j$. The assumption $g(\textbf{w}_{t+j},z_{t+j}) \approx g(\textbf{w}_{t},z_t^j)$ was made in synchronous SGD\cite{fbimagenet}. However, it often may not hold.
			
			If we denote $\tilde{\textbf{w}}_{t+1}^j = \textbf{w}_t-\hat{\eta} \frac{1}{M} \sum_{i< j} g(\textbf{w}_{t},z_t^i)$, we can unfold the summation in Eq.\ref{eq108} to  
			\begin{eqnarray}
			\tilde{\textbf{w}}_{t+1}^{j+1}=\tilde{\textbf{w}}_{t+1}^{j}-\hat{\eta}\frac{1}{M}g(\textbf{w}_t,z_t^j),j<M,\label{eq95}
			\end{eqnarray}then we have $\hat{\textbf{w}}_{t+1}=\tilde{\textbf{w}}_{t+1}^{M}$.
			We propose to use Eq.(5) in the main paper to compensate this assumption and apply delay-compensated gradient to update Eq.\ref{eq95} with:
			{\begin{eqnarray}
				&&g(\textbf{w}_{t+j},z_{t+j})\approx \tilde g(\tilde{\textbf{w}}_{t+1}^{j}, z_t^j) :=  g(\textbf{w}_t,z_t^j)+ \lambda g(\textbf{w}_t,z_t^j)\odot g(\textbf{w}_t,z_t^j) \odot (\tilde{\textbf{w}}_{t+1}^{j}-\textbf{w}_t) \big), \\
				&&\tilde{\textbf{w}}_{t+1}^{j+1}=\tilde{\textbf{w}}_{t+1}^{j}-\hat{\eta}\frac{1}{M}\tilde g(\tilde{\textbf{w}}_{t+1}^{j}, z_t^j), j<M.\label{eq97}
				\end{eqnarray}}Please note that we redefine the previous $\tilde{\textbf{w}}_{t+1}^{j+1}$ in Eq.\ref{eq97}. For $j>1$, we need to design an order to make $\tilde{\textbf{w}}_{t+1}^{j}\approx\textbf{w}_{t+j}$. Choosing $\tilde{\textbf{w}}_{t+1}^{j}$ according to the increasing order of $\|\tilde{\textbf{w}}_{t+1}^{j}-\textbf{w}_t\|^2$ can be used since the smaller distance with $\textbf{w}_t$ will induce more accurate approximation by using Taylor expansion.

\end{document}